\documentclass[msom,nonblindrev]{informs3}

\OneAndAHalfSpacedXII 

\usepackage{natbib}
 \bibpunct[, ]{(}{)}{,}{a}{}{,}%
 \def\bibfont{\small}%
\TheoremsNumberedThrough     

\EquationsNumberedThrough    

\usepackage{multirow}
\usepackage{caption}
\usepackage{subcaption}
\usepackage{enumitem}

\usepackage{xr-hyper}
\usepackage[hidelinks]{hyperref}
\usepackage{booktabs} 
\usepackage{placeins} 

\begin{document}


\RUNTITLE{Patient Outcome Predictions Improve Operations at a Large Hospital Network
}
\TITLE{Patient Outcome Predictions Improve Operations at a Large Hospital Network}

\ARTICLEAUTHORS{%
\AUTHOR{Liangyuan Na, Kimberly Villalobos Carballo}
\AFF{Operations Research Center, Massachusetts Institute of Technology, Cambridge, MA, 02139, USA, \EMAIL{\{lyna,kimvc\}@mit.edu}}
\AUTHOR{Jean Pauphilet}
\AFF{Management Science and Operations, London Business School, London, UK, \EMAIL{jpauphilet@london.edu}}
\AUTHOR{Ali Haddad-Sisakht}
\AFF{Dynamic Ideas LLC, Waltham, MA, 02452, USA, \EMAIL{ali@dynideas.com}}
\AUTHOR{Daniel Kombert, Melissa Boisjoli-Langlois, Andrew Castiglione, Maram Khalifa, Pooja Hebbal, Barry Stein}
\AFF{Hartford HealthCare, Hartford, CT, 06103, USA, \EMAIL{\{Daniel.Kombert, Melissa.Boisjoli-Langlois, Andrew.Castiglione, Maram.Khalifa, Pooja.Hebbal, Barry.Stein\}@hhchealth.org}}
\AUTHOR{Dimitris Bertsimas}
\AFF{Sloan School of Management, Massachusetts Institute of Technology, Cambridge, MA, 02139, USA, \EMAIL{dbertsim@mit.edu}}}

\ABSTRACT{
\textbf{\textit{Problem definition:}} 
Access to accurate predictions of patients’ outcomes can enhance medical staff's decision-making, which ultimately benefits all stakeholders in the hospitals. A large hospital network in the US has been collaborating with academics and consultants to predict short-term and long-term outcomes for all inpatients across their seven hospitals. \\
\textbf{\textit{Methodology/results:}} 
We develop machine learning models that predict the probabilities of next 24-hr/48-hr discharge and intensive care unit transfers, end-of-stay mortality and discharge dispositions. 
All models achieve high out-of-sample AUC ($75.7\%$--$92.5\%$) and are well calibrated. In addition, combining 48-hr discharge predictions with doctors' predictions simultaneously enables more patient discharges (10\%--28.7\%) and fewer 7-day/30-day readmissions ($p$-value $<0.001$). We implement an automated pipeline that extracts data and updates predictions every morning, as well as user-friendly software and a color-coded alert system to communicate these patient-level predictions (alongside explanations) to clinical teams. \\
\textbf{\textit{Managerial implications:}} Since we have been gradually deploying the tool, and training medical staff, over 200 doctors, nurses, and case managers across seven hospitals use it in their daily patient review process. We observe a significant reduction in the average length of stay (0.67 days per patient) following its adoption and anticipate substantial financial benefits  (between \$55 and \$72 million annually) for the healthcare system.
}

\HISTORY{}

\maketitle

\section{Introduction} \label{sec:intro}
The collection of patient-level information from Electronic Medical Records (EMRs) combined with advances in machine learning (ML) methodology creates opportunities to enhance hospital operations and clinical decision-making, especially for inpatients, who constitute a major part of hospital activities and revenues.

There is a recent, rich, and growing academic literature on machine learning models trained to predict inpatient outcomes.  
However, only a limited number of them are deployed in practice and make an impact. 
Since 2020, we, a collaboration of healthcare providers, academics, and data consultants, have been working together to develop predictive models on eight operational metrics for seven different hospitals (covering over 2,400 inpatient beds), to deploy these models in the entire hospital network, and to measure their impact on the operational performance of the hospitals. 

\subsection{The Hospital Network} \label{ssec:hhc}
The Hospital Network is the largest and most comprehensive healthcare network in Connecticut.
With 36,000 employees, the network operates in over 400 locations, including seven acute care hospitals, several behavioral health, physical therapy, and rehabilitation facilities, a multi-specialty physician group, a clinical care organization, services of regional home care and senior care, and a mobile
neighborhood health program.
In 2021, the system covered 104,696 transitions from inpatient care (i.e., movement to another healthcare setting), 555,358 patient days of hospitalization, and 406,949 emergency department visits, generating an operating revenue of \$5 billion.
The unified network provides a high standard of care with enhanced access, affordability, and equity in crucial specialties and institutes.
The network contains seven diverse hospitals, ranging from Hospital A (HA), one of the largest teaching hospitals in New England (867 licensed beds) to smaller (150 beds) community hospitals. 
Appendix section \ref{sec:a.hhc} and Table \ref{tab:hhc_stats} present in more detail the seven hospitals of the network.

\subsection{The Collaboration} \label{ssec:collab}
Over the past decade, the hospital network has been closely collaborating with our academic research group to improve decision-making in various parts of the healthcare system. 
The success and positive impact of over 20 analytics-based projects together in different parts of the hospital network built the necessary trust to envision a larger network-wide effort to improve operations using analytics.
For a project of this scale, we also partnered with a data consultancy company with deep expertise in the implementation of operations research and analytics tools.

\subsection{Summary of Contributions} \label{ssec:contributions}
First, we build an end-to-end ML pipeline, from data extraction and processing to a user-friendly software interface, and apply it to the seven hospitals. 
Using comprehensive data from patient EMRs, including patient status, clinical measurements, and laboratory results, we train ML models
~\citep[XGBoost,][]{chen2015xgboost} to daily predict eight inpatient outcomes: mortality risk, probability of discharge in the next 24 and 48 hours, discharge disposition, and intensive care unit (ICU) risk in the next 24 and 48 hours.
Our models achieve state-of-the-art accuracy ($75.7\%$--$92.5\%$ area under the receiver operating curve, i.e., AUC across all tasks and all hospitals) and are well calibrated. 
Moreover, we demonstrate that integrating our discharge predictions into physicians' decision-making process can identify more discharge opportunities with higher accuracy and lower readmission risk. 

Second, through multiple iterations with clinicians, we develop a software tool for doctors, nurses, and case managers, to integrate these predictions into their daily workflow. In addition to presenting raw risk scores, the tool provides patient-level explanations for each of the predictions as well as a color-coded alert system to help them quickly identify at-risk patients 
 (red alert) and potential imminent discharges (green alert). The features of our software help increase trust, clinical adoption, and easy integration into the patient progression rounds. 

Finally, we have been gradually deploying our solution in the hospitals, training physicians and nurses, and measuring its operational impact. 
As of April 2023, our tool is benefiting over 200 users and supporting daily operational decisions at the seven hospitals. 
We observe a significant reduction in the average length of stay (LOS) by 0.67 days per patient from our solution and project annual revenue uplift of \$55-\$72 million from our deployment in the system. 

\subsection{Our Tool in Action} \label{ssec:action} 
\begin{figure}[tbp]
    \centering
    \includegraphics[width=\textwidth]{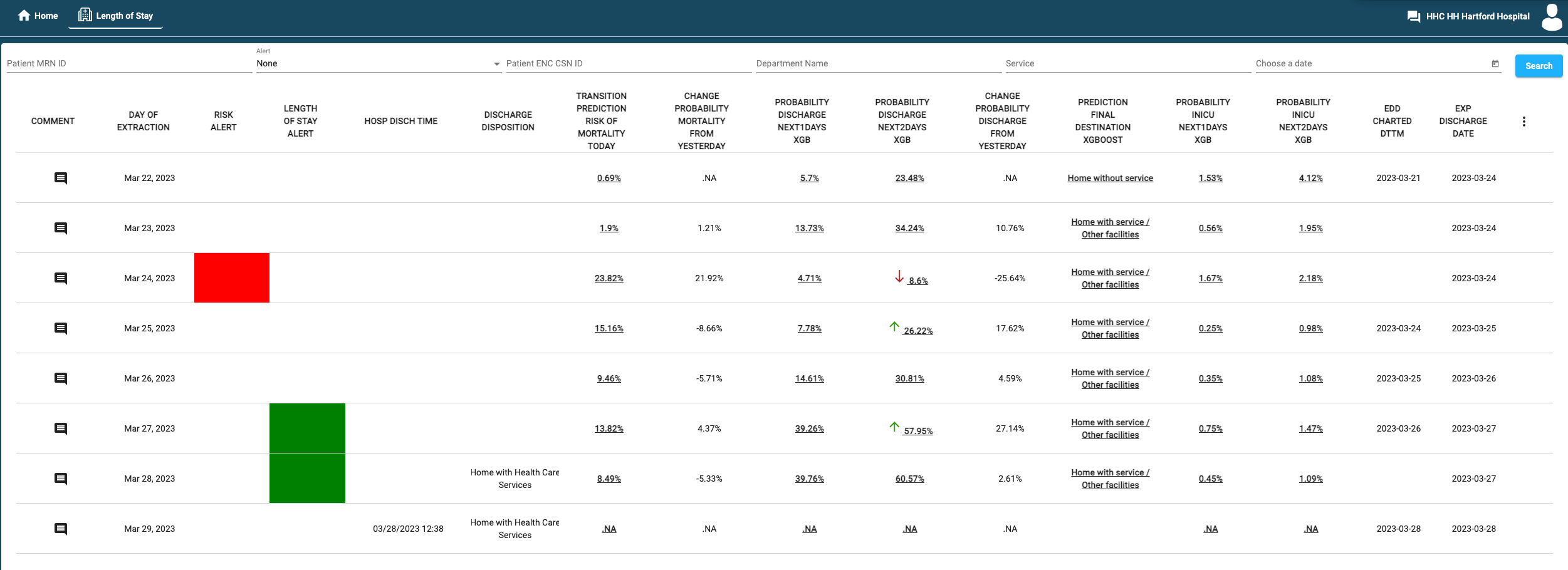}
    \caption{Trajectory of Predictions for an Example Patient.}
    \label{fig:pattraj}
\end{figure}

We now illustrate how our tool works on one patient trajectory. Figure~\ref{fig:pattraj} is a screenshot of our software solution for a particular patient. Each row corresponds to a day.
The patient is admitted on March 22, 2023. On the first day of the stay, the patient is assigned low probabilities of mortality and discharge.
Two days after admission, the condition of the patient deteriorates: discharge probabilities decrease and mortality risk increases, peaking at 23.82\% on March 24.
The system detects an exacerbation of the patient's condition and raises a red alert, which calls the attention of the caring team.
Over the following days, the mortality risks decrease towards patient recovery, and the probabilities of discharge gradually increase until discharge.
The system correctly delivers green alerts on the last two days of the stay. 
In particular, the probability of discharge in the next 48 hours (resp. 24 hours) started exceeding 0.4 (resp. 0.25) on the day before (resp. the day of) discharge. 
On March 28, the patient is discharged to Home with Health Care Services, as correctly predicted by the discharge disposition prediction model (from the final destination column).

\begin{figure}[tbp]
 \centering
    \begin{subfigure}[b]{0.44\textwidth}
     \centering
     \includegraphics[width=\textwidth,height=5cm]{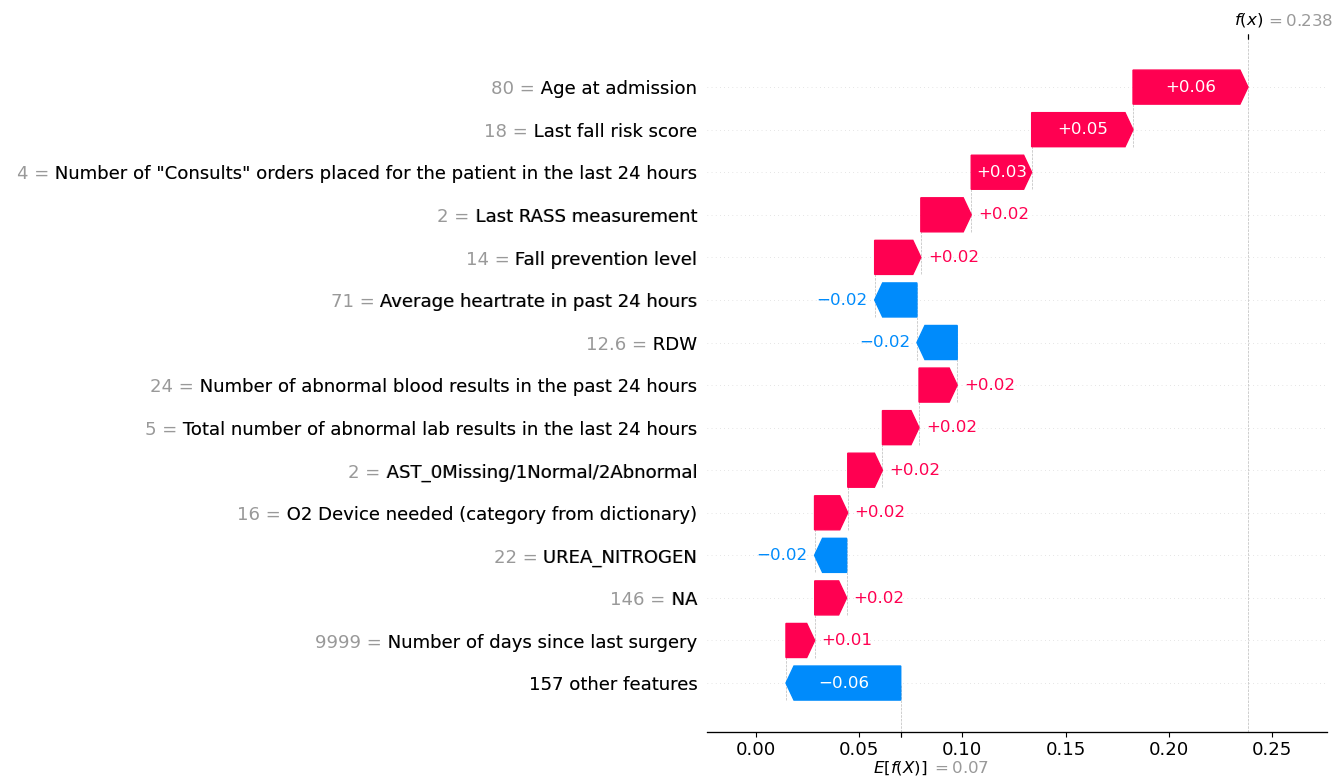}
     \subcaption{Mortality Risk on March 24.}
     \label{subfig:mortality}
 \end{subfigure}
 \hfill
 \begin{subfigure}[b]{0.55\textwidth}
     \centering
     \includegraphics[width=\textwidth,height=5cm]{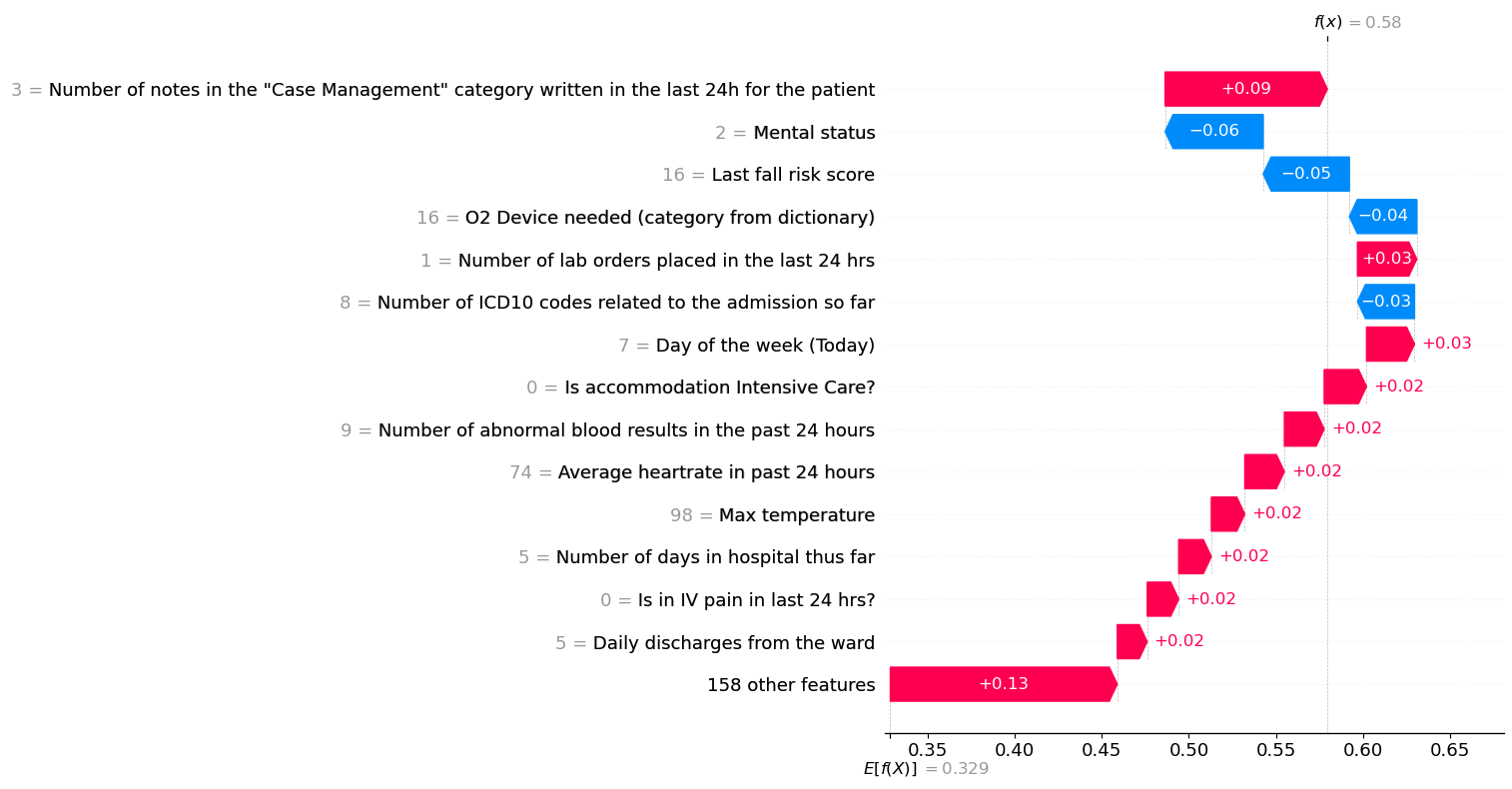}
  \subcaption{48-hr Discharge Probability on March 27.}
     \label{subfig:discharge}
 \end{subfigure}
    \caption{Prediction Explanation Plots for the Patient Example.}
    \label{fig:shap2}
\end{figure}

By clicking on each predicted value, the clinician can also access a waterfall plot to understand the patient-specific factors that explain such a prediction. 
Figure \ref{fig:shap2} displays these plots for the mortality prediction on March 24 (Figure \ref{subfig:mortality}) and the 48-hour discharge prediction on March 27 (Figure \ref{subfig:discharge}) for our patient. 
For a given risk score $f$, starting from the baseline risk in the population, $\mathbb{E}[f(X)]$ (here, the variable $X$ denotes all patient-level information used to make predictions), at the bottom of the plot, we add up the contributions of each variable to finally reach the patient-specific estimate, $f(x)$.
Figure~\ref{subfig:mortality}, for example, explains why the patient's mortality risk score on March 24 is 0.238, while the average prediction among all patients is 0.07.
The main factors explaining a higher-than-average prediction are the fact that the patient has a high age (+0.06), a high fall risk score assessment (+0.05), multiple consultation orders placed in the last 24 hours, and agitated status indicated by a Richmond Agitation Sedation Scale (RASS) measurement equal to 2 (+0.02), among others.
Meanwhile, these aspects are partially counterbalanced by several other variables that decrease the predicted mortality risk, including the average heart rate in the past 24 hours and red cell distribution width (RDW). 

\section{Problem Statement and Related Literature} \label{sec:litreview}
Anticipating short-term discharges informs about bed availability and can facilitate resource utilization while identifying discharge barriers provides clinical guidance to personalize the delivery of care. 
Furthermore, detecting patients with high mortality or ICU risk (or changes thereof) can alert the medical team and call their attention to those who need it the most. 
More broadly, discharge is a complex process involving the coordination of different stakeholders and resources, which could be anticipated given accurate predictions on discharge disposition early in a patient's stay. 

Accordingly, we focus our attention on the following patient outcomes: length of stay (LOS), mortality risk, discharge disposition, and ICU risk. There is a collection of models for such inpatient flow predictions in the literature. We refer to \citet{awad2017patient} for a comprehensive survey on mortality and LOS predictions and to \citet{mees2016predicting} for discharge destination predictions.

\subsection{Length of Stay} 
Models in the literature predict a variety of LOS related outcomes, such as daily discharge volume~\citep{zhu2015time}, next 24-hour discharge~\citep{safavi2019development, bertsimas2021predicting}, long LOS~\citep{bardak2021improving, bertsimas2021predicting}, and remaining LOS~\citep{wang2022high}.
Since our priority is prompt discharge identification, the medical team suggests predicting whether each patient will be discharged without expiration (i.e., without death) in the next 24 and 48 hours.
We note that patient death or transition to hospice are not considered as discharges. 
Predicting the next 48h discharges is critical and particularly useful for the hospitals since it helps physicians identify and prioritize patients who are ready for discharge while giving case management teams enough time to accelerate discharge preparations, which ultimately reduces patient burdens and direct operating costs in healthcare systems.

\subsection{Mortality Risk} Preventing death is one of the major responsibilities of hospitals.  
Physicians examine and evaluate patient daily charts with clinical knowledge and experience, but they encounter challenges in simultaneously processing hundreds of measurements and giving quick dynamic assessments for all patients.
We build models to help predict each patient's mortality risk, defined as the probability of each patient expiring or going to hospice at the end of the hospital stay. 
The high accuracies of mortality risk prediction models are demonstrated in previous works such as \cite{awad2017early,rajkomar2018scalable, jin2018improving,bardak2021improving}.
With such aid, the medical staff can give prioritization and rapid treatment plans to high-risk patients, which could potentially prevent their death. 
Moreover, the detection of increasing mortality risk over time alerts the care team of patients with worsening conditions, which can lead to more timely intervention and improve patient outcomes. 

\subsection{Discharge Disposition} It is also important to anticipate the patient's final destination after discharge. 
We divide the dispositions into three categories: home without service (i.e., discharged back home with self-care), expired or hospice (died at the hospital or transferred into hospice, where the latter is considered as near death), and home with service or other facilities (such as skilled nursing facility, rehab facility, long-term acute care hospital).
Differentiating discharge destinations early helps anticipate the need for post-discharge resources. 
In particular, the third discharge disposition requires additional case management efforts, such as contacting and obtaining approval from the care facility or ordering devices for service at home.
Combined with discharge time prediction, it can help the case management team coordinate post-discharge services and prevent logistical delays.

\subsection{ICU Risk} 
Uncertain intensive care demand and limited bed space make it particularly challenging to manage flows out of and into the ICUs, one of the most scarce and expensive resources in a hospital. 
Patients who need to enter an ICU and cannot do so promptly are often placed in secondary units, leading to higher readmission risk and extended LOS~\citep{kim2016association}.
On the other side, some patients are ready to leave the ICU but experience delays due to congestion in step-down units or delayed ICU discharge identification, which in turn congests the ICU, overflows other parts of the hospital~\citep{long2018boarding}, and prolongs boarding from the emergency department~\citep{mathews2018effect}. 
As exacerbated during the COVID-19 pandemic, predictions of ICU admission and mortality can help manage these valuable units~\citep{zhao2020prediction,covino2020predicting,subudhi2021comparing}.
We build models to predict the probability of being in the ICU in the next 24 and 48 hours, respectively. 
Specifically, we predict the probability of entering (resp. leaving) the ICU for patients currently not in (resp. currently in) the ICU.

Our objective, however, is not to predict for the sake of prediction but, more importantly, to implement our innovations in hospitals for real-world impact. Under such context, we aim to develop practical ML systems and deploy them in the entire hospital network. On this regard, our approach is similar to that of \cite{bertsimas2021predicting} who study a similar set of patient outcomes and integrate their predictions into hospital daily operations. In addition to using different ML models, their analysis, however, is single-center while we train, deploy, and evaluate the benefits of our models in different locations. Furthermore, the operational decisions and end-users are different. While they create a dashboard with unit-level predictions for the hospital flow management center, we work with doctors, nurses, and medical staff to leverage these patient-level predictions for the management of each patient individually.

\section{Methods} \label{sec:methods}
In this section, we describe our methodological approach, from data collection and feature engineering to predictive modeling and its integration as a decision-making support tool.

\subsection{Data Collection}
We first collect and build relevant data extracts for our various prediction tasks.
Since the hospital system uses a third-party EMR solution and since this is their first ML project applied to all inpatients in the network, there was no pre-existing pipeline for us to use, and thus we build the data extracts and pipeline from scratch. 

We start by replicating the pipeline of 
\citet{bertsimas2021predicting} and using the same set of variables as those they identify important, followed by weekly discussions with doctors, nurses, and the IT team to include other useful data sources and variables. 
Based on those discussions as well as data availability, we build 10 data extracts summarized in Table~\ref{tab:data_extract} in the Appendix.
The extracts provide information including demographics (e.g., age at admission), patient status (e.g., current service, oxygen device), clinical measurements (e.g., oxygen concentration, blood pressure), laboratory results (e.g., albumin, bilirubin), diagnoses, orders, procedures, notes, and others.

\begin{figure}[tbp]
    \centering
    \includegraphics[width=\textwidth]{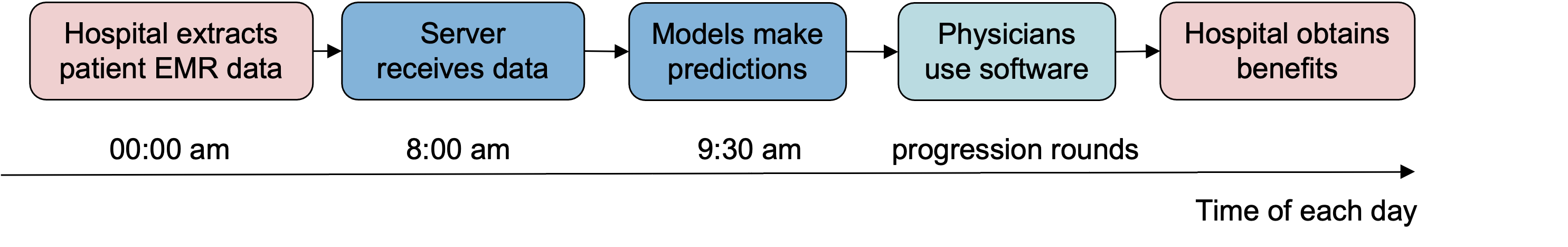}
    \caption{End-to-end Pipeline in Daily Production.}
    \label{fig:daily_pipeline}
\end{figure}

All the data is extracted from the EMRs directly, on the hospital network's IT system, password-protected, and then transferred to a dedicated database hosted on Amazon Web Service (AWS) machines by a Secure File Transfer Protocol server. 
Daily extracts are scheduled every day at midnight so that up-to-date data about current inpatients are received by 7:40 am every morning (see Figure~\ref{fig:daily_pipeline}).
EMRs were gradually deployed in the network so data availability for training purposes varies by hospital. Our database starts on 
January 2016 for Hospital A, January 2021 for Hospital F, and January 2020 
for the remaining five hospitals.
\subsection{Feature Curation}
Since our pipeline and predictions are updated daily, we curate a feature space where each row represents each patient day.
We create features from the following six groups.
1) Current conditions (e.g., department, whether in ICU); 
2) Lab results (e.g., albumin, white blood cell count); 
3) Clinical measurements (e.g., temperature, respiratory rate, heart rate); 
4) Time series summary statistics of operational variables (e.g., days in ICU); 
5) Patient information prior to current admission (e.g., age, previous admission); 
6) Auxiliary operational variables which are not patient-specific (e.g., day of the week).
Creating this collection of features requires a large amount of data processing, such as imputing missing data, parsing string formats, and encoding categorical variables.
A more comprehensive list of variables and details on data processing can be found in the \ref{sec:a.featureprocessing} Section of the Appendix.

\subsection{Machine Learning Modeling}

\noindent \textit{Inclusion-Exclusion Criteria:} We consider all inpatients during their hospitalization days.
Emergency Department patients and outpatients who are not admitted as an inpatient are excluded.
For each inpatient, we consider the range starting from their admission date until their discharge date.
We further filter the data depending on the prediction target.
We identify a set of special discharge dispositions: left against medical advice/AMA, still a patient, admitted as an inpatient, court/law Enforcement, ED dismiss/diverted Elsewhere.
For mortality and discharge disposition predictions, we exclude patients whose discharge disposition is missing because the target is missing, and exclude patients who have a special discharge disposition to reduce noise in the target.
For 24/48-hr discharge predictions, we exclude patients whose discharge disposition or discharge time is missing and exclude data points where the patient is discharged in the next 24/48 hours to one of the special dispositions.
For 24/48-hr ICU predictions, we include only patients who still are in the hospital in the next 24/48 hours.
Moreover, we include only patients who are currently in the ICU for leaving ICU predictions and include only patients currently not in the ICU for entering ICU predictions.
\vspace{0.2mm}

\noindent \textit{Split of Training, Validation, and Testing Sets:} For each hospital, we sort the data by record date and split it into 50\% training, 20\% validation, and 30\% testing set.
The train/validation/test set split is performed separately for each hospital and chronologically to reflect real-world implementation, where the models are trained on the past data and utilized for future prediction. 
Both imputation and machine learning models are trained on the training set, tuned on the validation set, and evaluated on the testing set. 
\vspace{0.2mm}

\noindent\textit{Prediction Models:} We consider a variety of machine learning models to make predictions.
Since patient predictions should be interpretable for medical staff, we started with interpretable ML models, including Optimal Classification Trees ~\citep[OCT,][]{bertsimas2017optimal} and sparse classification~\citep{bertsimas2017sparse}, implemented in the Interpretable AI software~\citep{InterpretableAI}.
OCT is a state-of-the-art interpretable decision tree model that divides the feature space using simple and understandable rules. 
The resulting OCT trees were closely examined and discussed by the doctors, which uncovered important insights as well as established doctors' initial understanding and trust in the algorithm.
As doctors got more receptive to using and understanding the models, we then shifted the objective to improve the performance of predictions.
We consider other machine learning models including XGBoost, LightGBM~\citep{ke2017lightgbm}, and Tabnet~\citep{arik2021tabnet}.
XGBoost is a gradient boosting method that ensembles a set of decision trees to make predictions in an additive fashion.
It consistently outperformed the other five methods across all prediction tasks on preliminary experiments so we decided to use it for our final models. 
To obtain higher accuracy, we also considered ensembling up to 10 XGBoost models together or creating more sophisticated features from time series measurement using the \texttt{tsfresh} package~\citep{christ2018time}, but decided that the marginal gain in accuracy did not justify the additional computational burden and lost of interpretability.
We train a separate model for each prediction task (a multi-class classification model for discharge disposition and binary classification models for the other target variables) and for each hospital.
We use the validation set to calibrate the hyper-parameters (depth of trees, learning rate, number of estimators, loss function, and L2 regularization rate).
\vspace{0.2mm}

\noindent \textit{Model Calibration:} For interpretability purposes, it is important for classification models to be well calibrated, i.e., that the numerical scores (scaled between 0 and 1) returned by the models correspond to the probability of the event of interest to occur. 
For example, if the predicted probability of discharging a patient in the next 48 hours is 0.2, a doctor will likely expect to interpret that this patient has a $20\%$ chance of being discharged. 
Note that calibration and accuracy are different issues (for example, dividing all scores by a factor of $2$ impacts calibration but keeps AUC constant). 
Therefore, we ensure that all our models are well calibrated by chronologically splitting the testing set into two halves, using the isotonic regression method \citep{zadrozny2002transforming} to calibrate the model on the first half, and then assessing the final calibration on the second half.
\vspace{0.2mm}

\noindent \textit{Computing Resources:} Data processing, feature engineering, model training, and offline testing are conducted in Python 3.5.2 with a parallelization strategy on a remote server with 32GB RAM Intel Xeon-P8 CPU per instance.
The online daily pipeline in production is run in Julia 1.5 and Python 3.8 on a cloud computing server via AWS with 64GB RAM and 8 CPUs.

\subsection{Predictive Analytics for Decision-Making}
To turn these predictive analytics into a decision support tool that is sustainably used by practitioners, we complement the raw probability predictions with an alert system and visual explanations for each prediction.
\vspace{0.2mm}

\noindent \textit{Color-coded Alert System:} It can be difficult for clinicians to quickly grasp the implications of a raw probability score and use it efficiently for decision-making. For instance, it is not obvious how to interpret a 0.28 risk of mortality or a 0.57 probability of discharge.
Moreover, medical staff are responsible for many patients, and they cannot read and process hundreds of probabilities including eight predictions for all their patients every day as part of their filled daily schedule. 
To overcome these challenges, we design an alert system to highlight the set of patients who are getting likely to be discharged and patients who are exacerbating with different colors. 
We send a green alert for patients that are ready for short-term discharge, i.e., whenever their probability of 24-hr or 48-hr discharge is above certain thresholds.
On the other hand, we send a red alert to warn patients who have a high risk or are exacerbating, i.e., if the mortality risk or the increase of mortality risk from the previous day reaches certain thresholds. 
\vspace{0.2mm}

\noindent \textit{Prediction Explanation Plot:} As doctors require sufficient clinical reasoning to make major decisions like patient discharge, it is critical to provide them with some interpretation of model predictions.
We compute Shapley values and SHapley Additive exPlanations~\citep[SHAP,][]{vstrumbelj2014explaining} on the XGBoost models to derive each feature's attribution to model predictions.
We use SHAP summary plots to visualize the top features of each model and their overall effect on the predictions, which is useful to audit the models and assess their validity with clinicians. 
We also produce SHAP plots at the individual level on every patient's prediction on each day, to provide a visual explanation of each predicted probability.

\section{Project Management Approach} \label{sec:project}
A major factor in the success of the large-scale deployment of an advanced analytics solution like this one lies in the way we organized the progression of the project over time and scheduled its extension from one to several hospitals. 

\subsection{Project Timeline}
\begin{figure}[tbp]
    \centering
    \includegraphics[width=\textwidth]{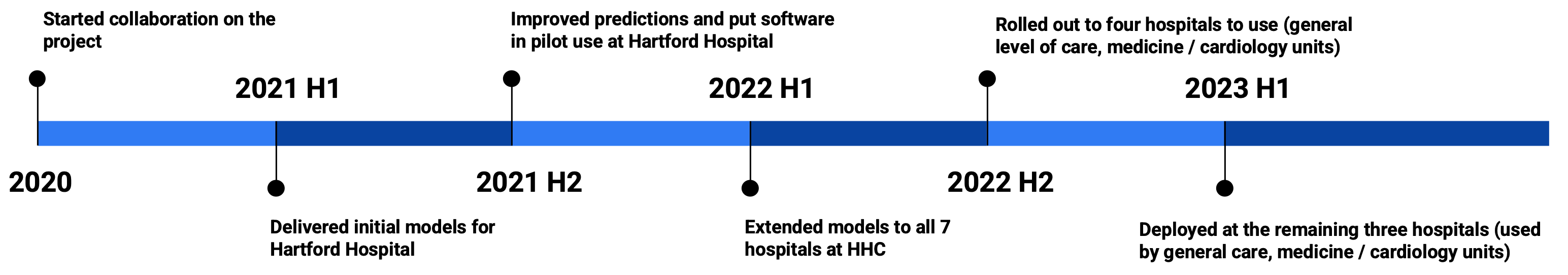}
    \caption{The Collaboration Timeline.}
    \label{fig:storyline}
\end{figure}
We have been collaborating on this solution since 2020.
Our first challenge was to interface with the third-party software the hospital system uses for managing all of their EMRs. In 2020, we worked on an automated pipeline to daily extract data from the EMR system to a dedicated database.  
We then focused our attention on HA, the main hospital and the largest hospital in the network, to develop a first proof of concept for our predictive models (including only a limited number of operational outcomes) and our doctor-facing interface. 
With this prototype, we constituted in 2021 a group of physician champions to serve as beta testers, collected feedback from their utilization and understanding of the models (regarding model accuracy, missing predictive information, additional relevant outcomes to predict, and the user interface), and iteratively improved the models and the software tool.
The tool was then rolled out for wider usage at HA in 2022.
Concurrently, we re-trained the models initially developed for HA to the other six hospitals and deployed them in production progressively between May 2022 and January 2023.
The timeline of the project is sketched in Figure~\ref{fig:storyline}.

\subsection{Pilot Implementation at Main Hospital} 
Instead of developing our models for all the hospitals simultaneously, we adopted a gradual approach and started with HA, the main hospital of the network. 
Since the second half of 2021, the main hospital tested the tool with four physician champions who are lead hospitalists of five medical units at HA, including two teaching units (BLISS 7 EAST and CONKLIN 4) and three non-teaching units (CENTER/NORTH 12 and CONKLIN 5).
Every week, the clinical and analytics teams met to review technical issues on the deployment of the model in production as well as to incorporate feedback from the physician leaders.

\begin{figure}[tbp]
    \centering
    \includegraphics[width=\textwidth]{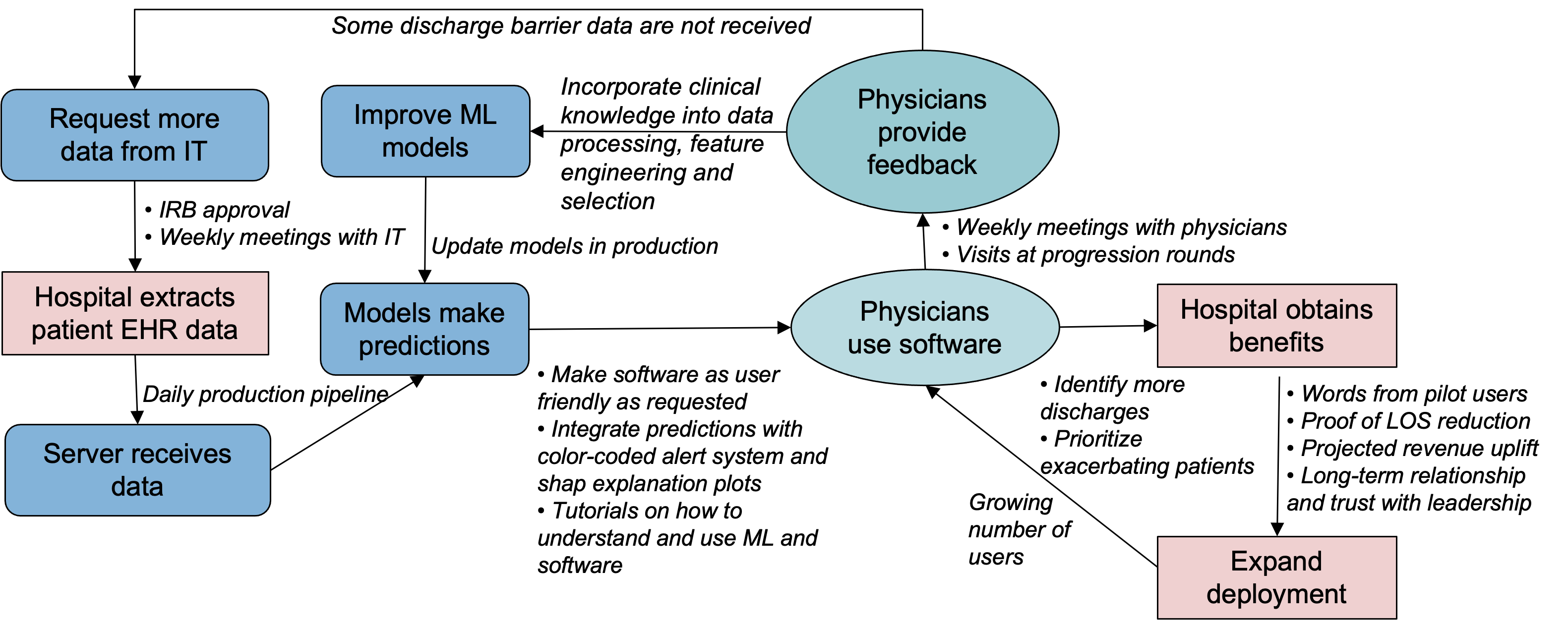}
    \caption{Implementation Feedback Loops.}
    \label{fig:feedback_loop}
\end{figure}

Figure~\ref{fig:feedback_loop} represents the different feedback loops between the technical team (blue rounded rectangle), the hospitals (red rectangle), and specific physicians (green oval).
Feedback from the physicians has been crucial to identifying other sources of data to integrate into our model, as well as defining the most relevant use cases for the prediction and improving the functionalities of our software tool. 
On a weekly basis, physicians reviewed patients and their predictions and provided written comments for each prediction. For example, when patients had a high predicted discharge probability but were not ready for discharge yet, physicians would note their current discharge barrier and the technical team would propose solutions to account for it. 
Through this process, the technical team incorporated physicians' clinical knowledge into the feature processing and engineering steps, e.g., by creating delta variables and time series summary statistics for some important features or by adjusting the missing value imputation based on medical knowledge.
Clinical expertise was also needed to decide the depth and breadth of the data extracts. For example, we initially received only one measurement of the RASS score per day (the latest) but later decided to extract all daily measurements to capture deterioration/improvement in the patient's anxiety level. 

Having a software interface to share the outputs of our models in a convenient, interpretable, and streamlined way with physicians significantly accelerated the adoption of the tool and its continuous improvement. The technical team further facilitated the integration of the tool by providing tutorials and lectures about machine learning to doctors at the hospitals.
We cannot emphasize enough how instrumental this pilot implementation with physician champions at HA has been in establishing a close relationship and trust between the analytics team and the doctors and nurses, and in enabling the wider roll-out of the solution.

\subsection{Scaling to Multiple Hospitals}
Only once built the data extraction pipelines and the ML models, and ran through a couple of iterations with teams at HA, did we start replicating this development (including the data extraction part) to other hospitals. 
After an evaluation of the impact of our solution on the pilot conducted at HA (presented in the \ref{sec:impact} Section), the hospital network decided to extend our work to 
three other hospitals---Hospital B (HB), Hospital C (HC), and Hospital D (HD)---chosen for the diverse set of activities they cover and for their appetite for experimentation.
Finally, we included the remaining three hospitals---Hospital E (HE), Hospital F (HF), and Hospital G (HG).

By extending our data extraction and processing to different institutions, we designed a fairly robust and generalizable feature processing pipeline (described in the 
\ref{sec:a.featureprocessing} Section of the Appendix). 
After feature curation and processing, we chose to train separate ML models for each hospital for different reasons.
First, as described in the \ref{sec:a.hhc} Section in the Appendix, the seven hospitals are very diverse in size, service, levels of care, and patient populations.
Second, developing and deploying models for a large hospital network requires a staggered roll-out. 
As opposed to developing a model for all hospitals at once, we developed and implemented them gradually to earn the trust and support of the leadership.
Third, since each hospital has its own encoder and imputer (see \ref{sec:a.featureprocessing} Section of the Appendix), we cannot apply one common model for all hospitals. In an early stage, we tried applying the model trained for HA directly to three other hospitals and, as expected, we obtained low performance due to different ways of encoding (e.g., for department and service).

\section{Results} \label{sec:results}
We present statistics, evaluation, and analysis of the models.
Unless specified otherwise, all results in this section are evaluated in the testing sets for each hospital. 

\subsection{ML Model Evaluation}

\noindent \textit{Accuracy:} The performance of the binary classification models is assessed by the AUC. 
For discharge disposition, multiclass AUCs are computed on each class against the rest.
Table~\ref{tab:auc} presents out-of-sample AUCs of eight prediction tasks for seven hospitals.
For hospitals HA--HF, AUCs range from 0.905--0.925 for mortality, 0.858--0.884 for discharge disposition, 0.812--0.848 for 24-hr discharge, 0.816--0.852 for 48-hr discharge, 0.850--0.872 for 24-hr entering ICU, 0.812--0.896 for 24-hr leaving ICU, 0.811--0.847 for 48-hr entering ICU, and 0.833--0.896 for 48-hr leaving ICU.
Our models achieve state-of-the-art performances compared with the literature described in the \ref{sec:litreview} Section.
Compared with the other six hospitals, HG has lower AUCs in mortality and discharge-related predictions, which are likely due to the smaller data size (see Table~\ref{tab:data_size} and Table~\ref{tab:target_size}) and the higher complexity of differentiating among less critical patients who have a much higher proportion of being discharged in the next 48 hours than other hospitals (see Table~\ref{tab:target_prop}).
A full summary statistics of the data at each hospital for each prediction task is reported in the \ref{ssec:a.stats} Section in the Appendix.

\begin{table}[tbp]
\centering
\caption{AUC Metrics for All Predictions in All 7 Hospitals.}
\label{tab:auc}
\setlength\extrarowheight{-10pt}
\begin{tabular}{llllllll}
Hospital Prediction   & HA    & HB    & HC    & HD  & HE   & HF    & HG      \\ \midrule
Mortality   & 0.919  & 0.915 & 0.902  & 0.905 & 0.925 & 0.925 & 0.888       \\ 
Discharge Disposition  & 0.884  & 0.858 & 0.871& 0.884 & 0.879 & 0.869 & 0.802       \\ 
Discharge 24 hr   & 0.857  & 0.832 & 0.812 & 0.844 & 0.837 & 0.848 & 0.757       \\ 
Discharge 48 hr  & 0.852  & 0.830 & 0.816  & 0.843 & 0.836 & 0.841 & 0.768       \\ 
Enter ICU 24 hr & 0.868  & 0.867 & 0.853  & 0.868 & 0.872 & 0.850 & \multirow{4}{*}{No ICU} \\ 
Leave ICU 24 hr  & 0.871 & 0.896 & 0.830  & 0.883 & 0.887 & 0.812 &     \\ 
Enter ICU 48 hr  & 0.818 & 0.834 & 0.813  & 0.811 & 0.847 & 0.820 &     \\ 
Leave ICU 48 hr   & 0.865 & 0.896 & 0.848 & 0.880 & 0.876 & 0.833 &     \\ 
\end{tabular}
\end{table}
\vspace{0.2mm}

\noindent \textit{Model Calibration:} 
As discussed in the \ref{sec:methods} Section, it is important for classification models to be well calibrated. Accordingly, we calibrate our models on the first half of the testing set and assess the calibration on the second half using calibration curves. Detailed assessment of the calibration of our models is presented in the \ref{ssec:a.calibration} Section of the Appendix.

\subsection{Alert System Assessment}
\begin{figure}[tbp]
 \centering
    \begin{subfigure}[b]{0.49\textwidth}
     \centering
     \includegraphics[scale=0.5]
     {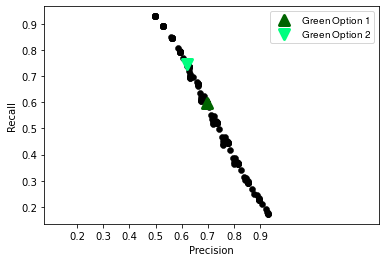}
     \caption{Green Alerts for Discharge.}
     \label{subfig:thr_green}
 \end{subfigure}
 \hfill
 \begin{subfigure}[b]{0.49\textwidth}
     \centering
     \includegraphics[scale=0.5]{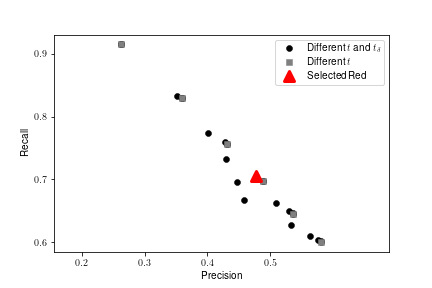}
  \caption{Red Alert for Mortality.}
     \label{subfig:thr_red}
 \end{subfigure}
\caption{Out-of-Sample Precision-Recall Curves under Different Thresholds for Colored Alert System at HA.}
     \label{subfig:thr}
\end{figure}

We select the thresholds for the color-coded alert system depending on the resulting precision/recall trade-off they provide for identifying discharges (green alert) and high-risk patients (red alert), as illustrated in Figure~\ref{subfig:thr} for HA.

For discharge prediction, our objective is to correctly mark (with a green alert) the patients that will be discharged in the next 48 hours. Precision represents the proportion of actual discharges among patients that are marked (also referred to as positive predicted value) and recall is the proportion of patients marked among all actual discharges (true positive rate). 
We raise an alert whenever the predicted probabilities of being discharged in the next 24 or in 48 hours exceed a threshold, $t_{24}$ and $t_{48}$ respectively. Figure~\ref{subfig:thr_green} represents the precision and recall for different threshold values, $t_{24}, t_{48} \in [0,1]$. 
We observe that the precision-recall relationship is almost linear. 
The hospital network wants to achieve a precision of around 0.7 for 48-hour discharge prediction, so we define option one of the green alerts as $t_{24} = t_{48} = 0.5$ 
(dark green upward triangle in the plot), yielding 0.698 precision and 0.598 recall.
After three months of utilization, the medical team expressed the need to identify more potential discharges, at the expense of lower precision.
Accordingly, we lowered the threshold for a green alert, defined as $t_{24} = 0.25$, $t_{48} = 0.4$ 
(light green downward triangle in the plot), which gives 0.621 precision and 0.746 recall.

For mortality prediction, our objective is to mark with a red alert two types of patients: those who will likely expire (i.e., death or hospice), and those who have worsening conditions. 
Accordingly, we raise an alert whenever the predicted mortality probability exceeds a threshold $t$, or when its absolute change compared with the previous day exceeds $t_\delta$.
Figure~\ref{subfig:thr_red} represents the precision/recall trade-off for $t \in [0.05,0.3], t_\delta = 1$ (gray squares) and for $t \in [0.05,0.3], t_\delta \in [0.05,0.3]$ (black circles). We observe that incorporating a criterion based on variations in mortality scores ($t_\delta < 1$) provides more granularity in terms of precision/recall trade-off, although it can lead to strictly worst performance if not calibrated carefully. Also, it spotlights patients whose condition is deteriorating, which the medical team can find even more helpful than predicting an absolute probability of mortality. 
For example, doctors find the tool more helpful in identifying a young patient's mortality risk increased from low probabilities on previous days, which would call for the provider team's action of interference, compared with reporting  an 80-year-old patient's consistently high mortality risk, which typically would have been already expected from their clinical assessment. 
Given that patient lives are at stake, doctors emphasize having a high recall and are less concerned about low precision (i.e., high false alarm rate). 
Therefore, we define red alerts as $t=0.2$ and $t_\delta  = 0.1$, which gives 0.477 precision and 0.705 recall.
    
We report the final accuracy, precision, and recall of our three alerts at all seven hospitals in the \ref{ssec:a.alert} Section of the Appendix.

\subsection{Assistance to Medical Staff}
To evaluate how our models can help the decision-making for doctors, nurses, and case management teams, we compare their accuracy with that of physicians. 
Every day at progression rounds, a team of attending physicians, residents, and nurses 
reviews each patient's chart information, discusses their case, and enters (or updates) an Expected Discharge Date (EDD) for each patient.
To compare with our models' outputs, we consider EDD as a score (by converting it into a time to discharge) to rank patients based on how soon they are likely to be discharged and we evaluate its AUC. 
In line with our outcome definition for the discharge prediction models, we compare the doctors' and models' discharge predictions on patients whose discharge dispositions are neither hospice nor expiration and report their respective AUC in Table~\ref{tab:comp_auc}.
We observe that our models achieve higher AUCs than the doctors for all hospitals and that the improvement is higher for 48-hr than for 24-hr predictions. 
The improvement ranges between 0.063--0.296 for 48-hr discharge and ranges between 0.024--0.283 for 24-hr discharge.

\begin{table}[tbp]
\centering
\caption{Comparison of AUC Metrics Between Doctors' and Models' Discharge Predictions.}
\label{tab:comp_auc}
\setlength\extrarowheight{-10pt}
\begin{tabular}{c|ccc|ccc}
\multicolumn{1}{c}{} & \multicolumn{3}{c}{48-hr Discharge AUC} & \multicolumn{3}{c}{24-hr Discharge AUC} \\\midrule
    Hospital & Doctor & Model    & Increment    & Doctor & Model    & Increment    \\ \midrule
HA       & 0.644              & 0.834 & 0.190        & 0.678              & 0.843 & 0.165       \\
HB       & 0.689              & 0.803 & 0.114       & 0.732              & 0.811 & 0.079       \\
HC       & 0.647              & 0.786 & 0.139       & 0.696              & 0.789 & 0.093       \\
HD     & 0.603              & 0.817 & 0.214       & 0.642              & 0.824 & 0.182       \\
HE      & 0.582              & 0.809 & 0.227       & 0.606              & 0.816 & 0.210        \\
HF       & 0.524              & 0.82  & 0.296       & 0.548              & 0.831 & 0.283       \\
HG       & 0.683              & 0.746 & 0.063       & 0.718              & 0.742 & 0.024          \\
\end{tabular}
\end{table}

We also investigate how doctors perform (in terms of precision and recall) at identifying 48-hour discharges compared with our green alert system. We report the precision and recall of doctors for each of the seven hospitals in Table \ref{tab:comp_prec}. In comparison with the green alert (see, e.g., Table \ref{tab:sel_prec_rec}), doctors generally demonstrate lower precision but higher recall, thus predicting more discharges than the green alerts. 
Greater benefits can be obtained by combining the two predictions together. 
In particular, if we predict a discharge when both the doctor's EDD is less than 48 hours away and a green alert is raised (`Doctor \texttt{AND} Green' column in Table \ref{tab:comp_prec}), we increase precision by 0.121--0.294 compared with doctors'. 
Furthermore, we can achieve higher recall by predicting a discharge whenever the doctor predicts discharge or a green alert is raised (`Doctor \texttt{OR} Green' column). By doing so, we complement doctors' input to identify potential discharge cases that doctors would otherwise have missed and increase the recall of doctors by 0.1--0.287 across seven hospitals.
These results suggest that by referencing the green alert system, doctors, nurses, and the case management team can identify more discharges and do so more precisely, which can further reduce patients' LOS.

\begin{table}[tbp]
\centering
\setlength\extrarowheight{-10pt}
\caption{Precision and Recall Improvement of Doctors' 48=h Discharge Predictions by Referencing Green Alert.}
\label{tab:comp_prec}\begin{tabular}{c|ccc|ccc}
\multicolumn{1}{c}{}    & \multicolumn{3}{c}{Precision}                    & \multicolumn{3}{c}{Recall}                 \\ \midrule
Hospital & Doctor & Doctor \texttt{AND} Green & Increment & Doctor & Doctor \texttt{OR} Green & Increment \\\midrule
HA       & 0.500  & 0.720                     & 0.220       & 0.681  & 0.878               & 0.197       \\
HB       & 0.522  & 0.713                     & 0.191       & 0.775  & 0.917               & 0.143       \\
HC       & 0.510  & 0.671                     & 0.161       & 0.792  & 0.926               & 0.134       \\
HD     & 0.478  & 0.691                     & 0.213       & 0.656  & 0.882               & 0.226       \\
HE      & 0.510  & 0.705                     & 0.195       & 0.605  & 0.892               & 0.287       \\
HF       & 0.411  & 0.705                     & 0.294       & 0.547  & 0.824               & 0.277       \\
HG       & 0.596  & 0.718                     & 0.121       & 0.806  & 0.906               & 0.100      \\ 
\end{tabular}
\end{table}

\subsection{Representation of Readmission Risk}
In addition to accurate, timely discharges, it is also important to ensure the safety of the discharge. 
According to Medicare, around 20\% of hospital-discharged patients are readmitted within 30 days, resulting in billions of dollars in annual costs~\citep{jencks2009rehospitalizations}. 
Early readmissions, which occur within 7 days post-discharge, are particularly closely linked to premature discharges, where a tradeoff exists between LOS and readmission~\citep{koekkoek2011hospitalists}. 
As predicting post-discharge outcomes can be challenging due to the many complex factors involved, models predict 30-day and 7-day readmission with moderate discriminative ability~\citep{zhou2016utility,saleh2020can}. 
Although our models for patient outcomes up to and including discharge do not consider readmission or other post-discharge outcomes, we are interested in exploring their potential associations. 
Specifically, we examine the relationship between our discharge prediction tool and the risk of patient readmission within 7 days and 30 days, respectively. 
We evaluate our findings using data from January to April 2022, which represents an out-of-sample period in the testing sets for all 7 hospitals prior to the models being put into production.

\begin{figure}[tbp]
 \centering
    \begin{subfigure}[b]{0.49\textwidth}
     \centering
     \includegraphics[width=\textwidth,height=5cm]{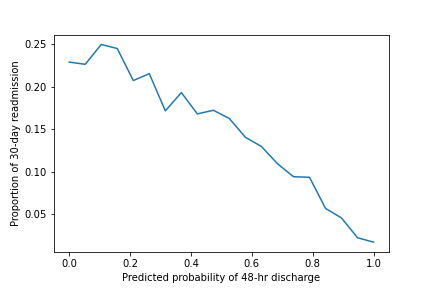}
 \end{subfigure}
 \hfill
 \begin{subfigure}[b]{0.49\textwidth}
     \centering
     \includegraphics[width=\textwidth,height=5cm]{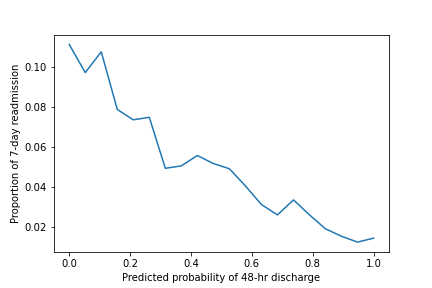}
 \end{subfigure}
    \caption{Readmission Risk in the Next 30 (left panel) and 7 (right panel) Days vs 48-hour Discharge Probability.}
    \label{fig:readmission_bucket}
\end{figure}

We begin by discretizing the probability of discharge within the next 48 hours into fixed buckets with 5\% probability intervals. 
For each bucket, we compute the proportions of 30-day readmission and 7-day readmission among patients within that probability bucket and who get discharged within the next 48 hours. 
As shown in Figure~\ref{fig:readmission_bucket}, we observe a negative, almost linear, relationship between the two. 
Specifically, higher predicted discharge probabilities correspond to lower readmission risks for both 30-day and 7-day readmissions. 
This suggests that our model's confidence in a patient's likelihood of being discharged within 48 hours is also indicative of the safety of the discharge. Importantly, this relationship holds even though readmission outcomes are not included as inputs or outputs during the training of our discharge prediction models.

\begin{figure}[tbp]
     \centering
     \includegraphics[width=0.8\textwidth]{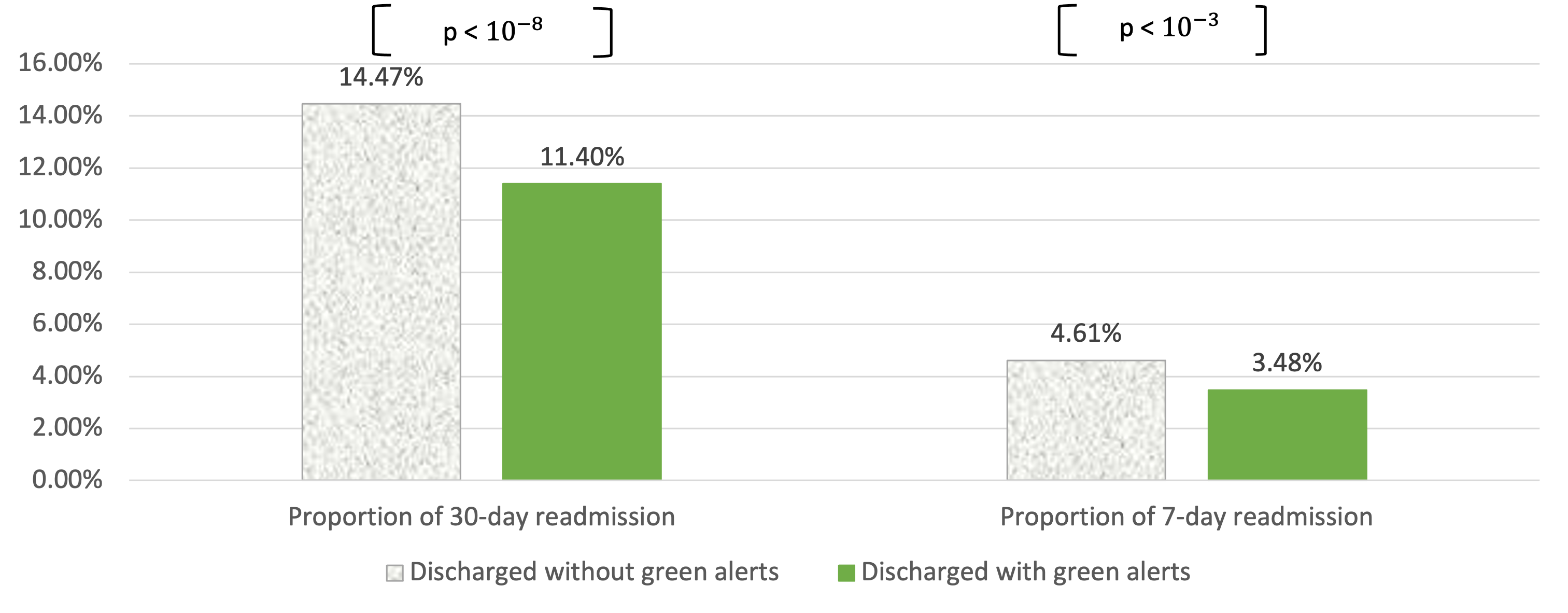}
    \caption{Readmission Risks for Patients Discharged without vs with Green Alerts.}
    \label{fig:readmission_mean_diff}
\end{figure}

We also investigate the relationship between our green alert system and readmission risk. Specifically, we compare the proportions of patients who are subsequently readmitted within 30 days or 7 days, among those discharged with and without a green alert, respectively.  
As shown in Figure~\ref{fig:readmission_mean_diff}, patients who receive a green alert 48 hours before discharge have 3.07\% lower 30-day readmission risk and 1.13\% lower 7-day readmission risk compared to those discharged without a green alert. 
One-sided Welch’s t-tests with different variance groups reject null hypotheses  with p-value $< 10^{-8}$ for 30-day and p-value $< 10^{-3}$ for 7-day, concluding that patients discharged with green alerts have statistically significantly lower readmission rates than those discharged without green alerts. 
Odds ratios for 30-day and 7-day readmissions for patients without a green alert compared to those with are 1.32 and 1.34, respectively. 
This indicates that the odds of being readmitted within 30 days and 7 days are increased each by over 30\% for patients who do not receive a green alert. 
These findings suggest an additional potential use of our green alert system to screen readmission risks and advise more confidence discharges. 

\subsection{Understanding Predictions}
We show SHAP summary plots of example prediction models on sample hospitals to understand how the top 30 features of each model impact the prediction in the testing set.
In the plots, positive SHAP values represent an increase in predicted probability whereas negative values represent a decrease; the colors ranging from red to blue represent high to low values of the features.

\begin{figure}[tbp]
 \centering
    \begin{subfigure}[b]{0.49\textwidth}
     \centering
     \includegraphics[width=\textwidth,height=5cm]{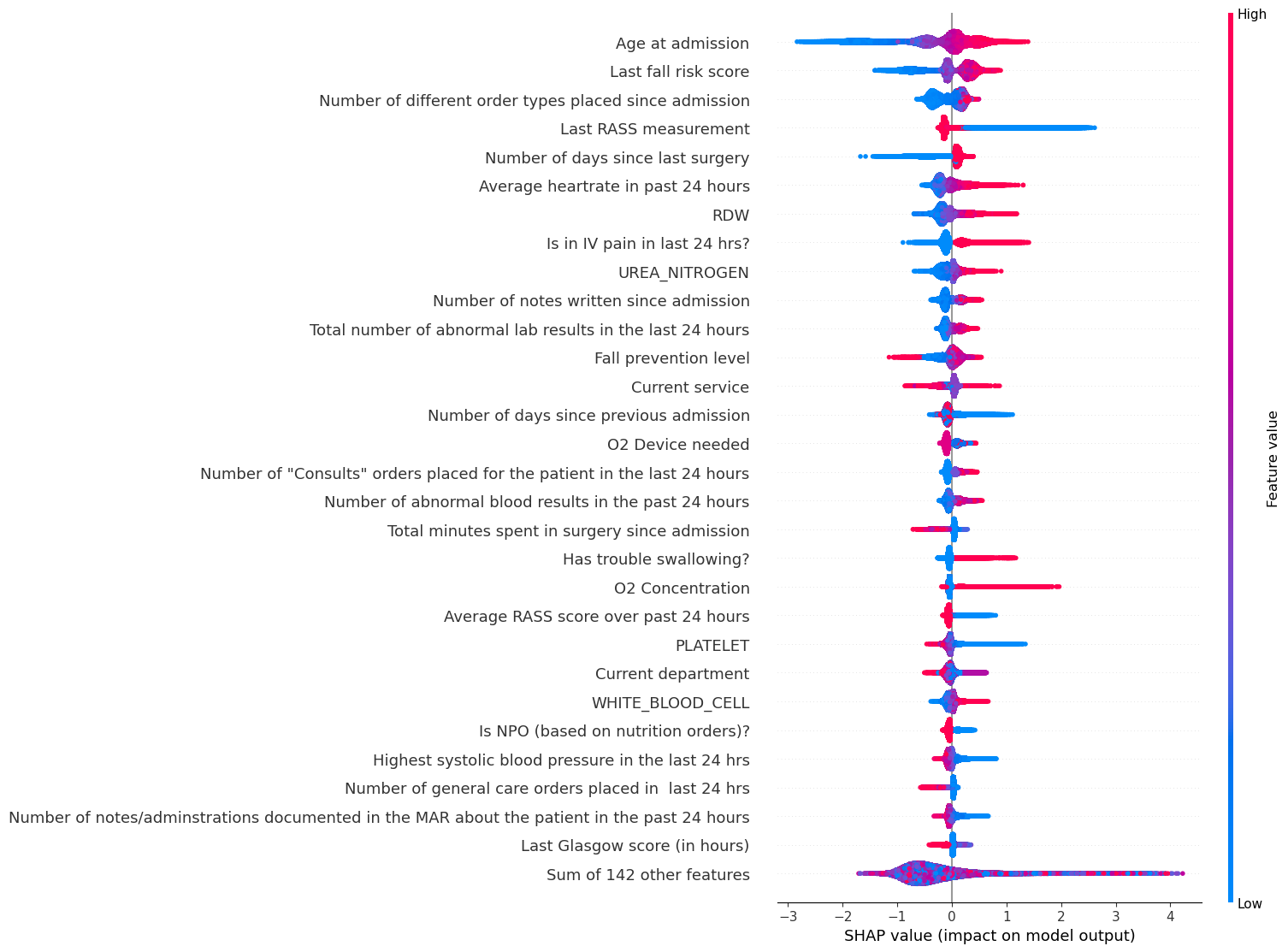}
     \caption{HA Mortality.}
     \label{subfig:hhmortality}
 \end{subfigure}
 \hfill
 \begin{subfigure}[b]{0.49\textwidth}
     \centering
     \includegraphics[width=\textwidth,height=5cm]{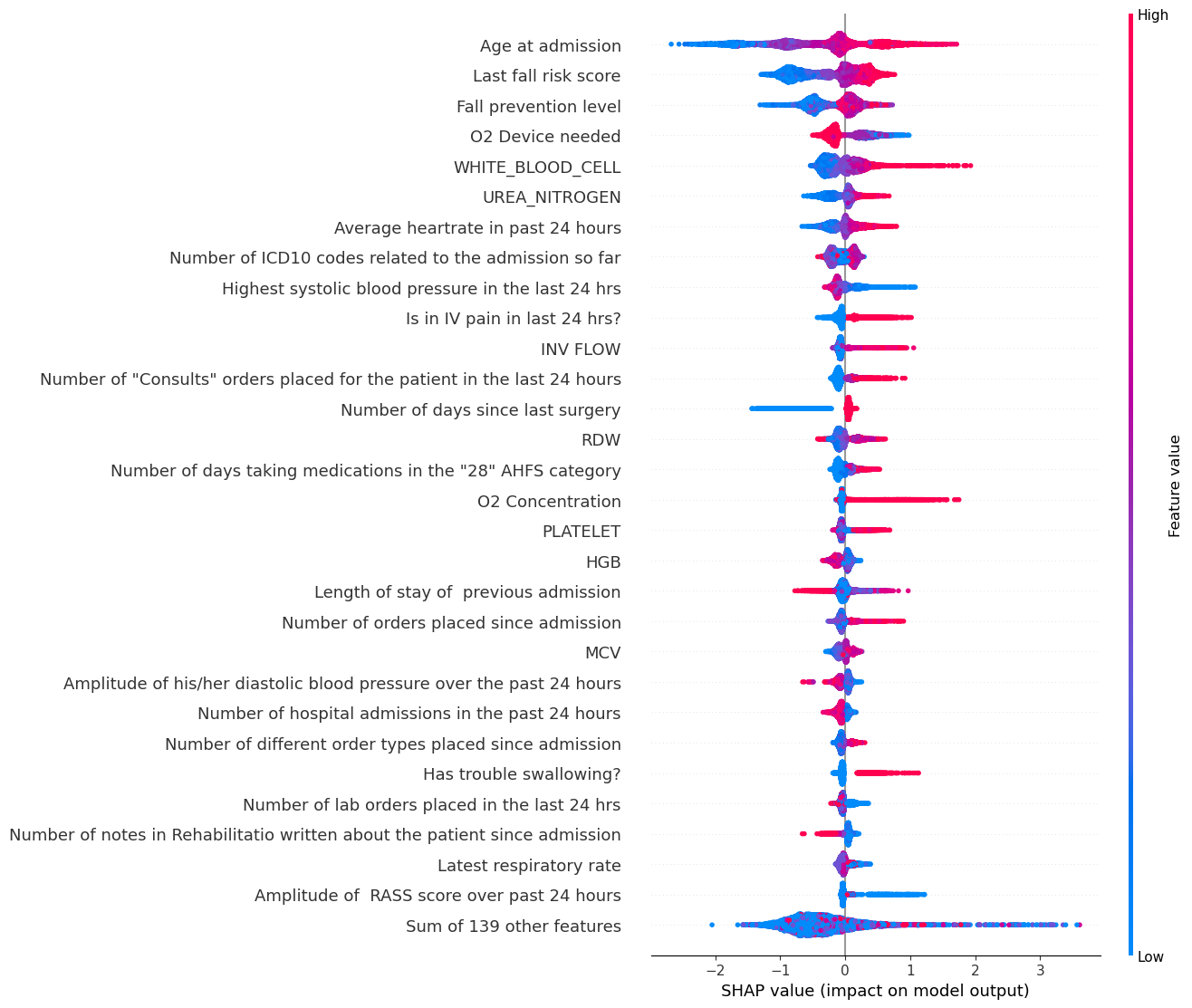}
  \caption{HG Mortality.}
     \label{subfig:whmortality}
 \end{subfigure}
    \caption{SHAP Summary Plots for Mortality Predictions.}
    \label{fig:shap}
\end{figure}

We analyze Figure~\ref{fig:shap} for mortality models.
For some important variables, such as age, fall risk score, heart rate, RDW, intravenous (IV) pain, and white blood cells, higher values drive higher mortality risks.
Other important variables include the number of order types since admission, the number of days since the last surgery, the oxygen (O2) device, fall prevention level, and more.
While most of these variables align with doctors' experiences and intuitions, others differ from what doctors typically use to make their assessments.
For example, according to our model, RDW plays a major role in mortality predictions; however, this feature was not one of the main variables doctors considered during their assessment. After some literature review, we learned that multiple papers found a similar significance of RDW in mortality predictions~\citep{csenol2013red, wang2018red, soni2021significance}, and following careful examination doctors agreed to incorporate RDW in their mortality assessment. This anecdote illustrates how interpretable machine learning can help doctors learn and enhance their clinical evaluation.

A comparison of the HA model (Figure~\ref{subfig:hhmortality}) and the HG model (Figure~\ref{subfig:whmortality}) shows that while the majority of the important variables are the same between the two hospitals, a small number of differences exists. 
For instance, the current department is a crucial factor in the HA model's predictions, but it does not rank among the top 30 features in the HG model. 
Reviews with doctors suggest this disparity might be due to the fact that HA has a broad range of departments catering to a diverse patient population, with varying levels of care and specialties, whereas HG has a small collection of departments that do not include critical or intensive care for a more homogeneous patient group. 
Further, we present additional SHAP plots on discharge and ICU predictions in Figure~\ref{fig:shap_icu} and analyze them in Section \ref{ssec:a.shap} of the Appendix.

\section{Implementation and Impact} \label{sec:impact}
As presented in the \ref{sec:project} Section, the solution was first piloted with physicians at HA. This pilot phase helped refine the user interface and conduct a first impact evaluation, before extending the deployment to the other hospitals. 

\subsection{Software Implementation}
\begin{figure}[tbp]
    \centering
    \includegraphics[width=\textwidth]{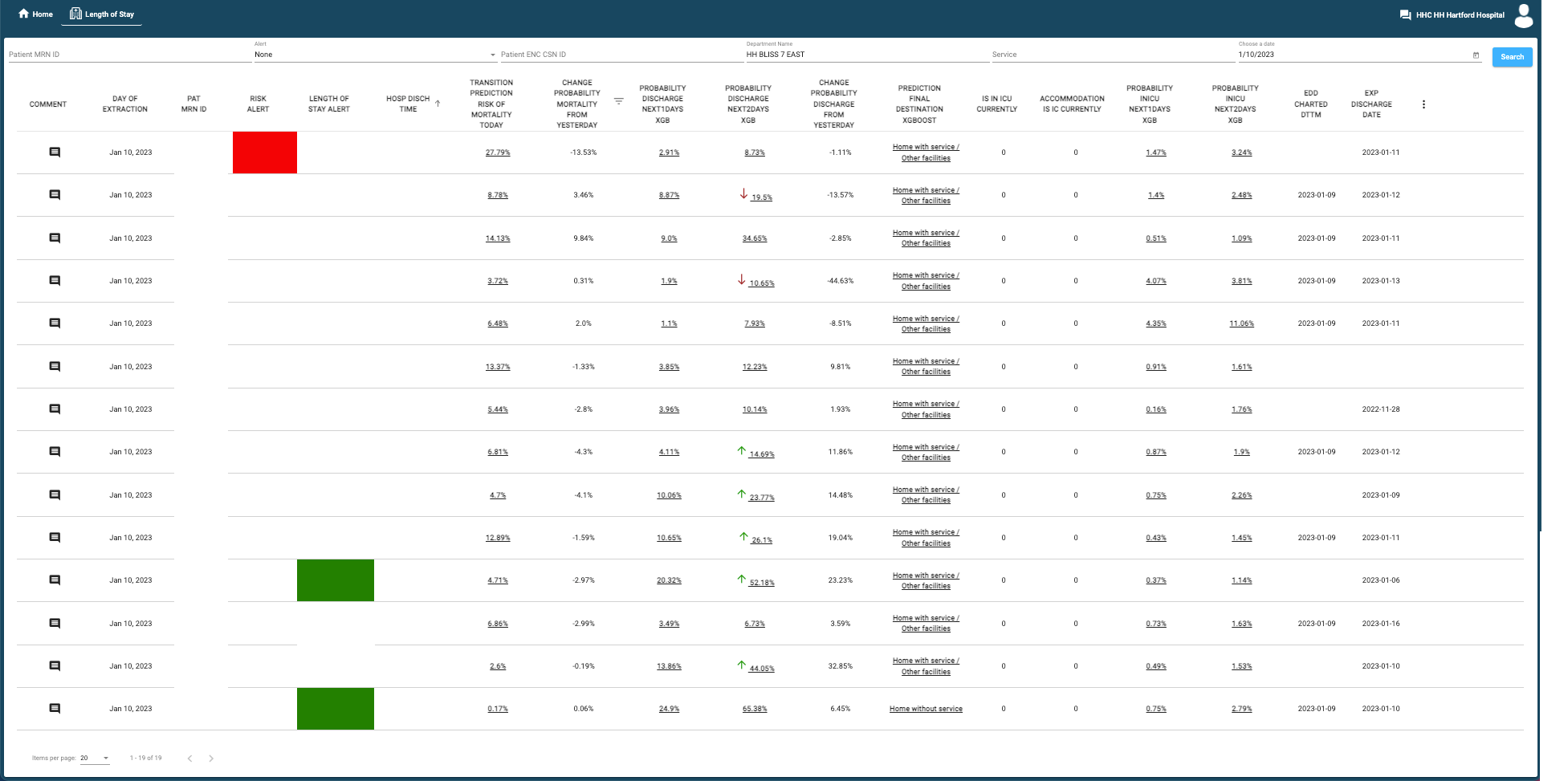}
    \caption{Table of predictions (with patient IDs de-identified).}
    \label{fig:los_table}
\end{figure}

Each doctor or medical staff can log into their account with two-factor authentication, select their hospital, and enter the application.
In the tool, patients can be filtered by a variety of features (e.g., department, date, alert, patient ID). For example, Figure~\ref{fig:los_table}  shows
a list of patients and their associated predictions in the HA Bliss 7 East unit on January 10th, 2023.
Each row corresponds to one patient on a given day.
Each user can customize their interface and the subset of columns to display among the following five categories:
1) Basic information, e.g., current date, patient ID, current department, room, bed number, service, and level of care; 
2) Predicted probabilities for the eight patient operational characteristics on the current day; 
3) The predicted probabilities from the previous day or the change in these probabilities; 
    An arrow $\textcolor{green}{\uparrow}$ (resp. $\textcolor{red}{\downarrow}$) is prefixed if the 48-hour discharge probability increases (resp. decreases) by at least 0.1 compared to the previous day.
4) Color-coded (green and red)  alerts; 
5) Doctor's expected discharge date with charted time. 
The software also supports printing and can be accessed from mobile devices. 
Upon clicking the comment icon, the user can provide feedback on the prediction for the team to review. 

As discussed in the \ref{sec:intro} Section, the clinician has also access to a SHAP waterfall plot for each patient prediction (upon clicking on the predicted value) to understand the patient-specific factors that explain the prediction. 
Figure \ref{fig:shap2} displays two examples of SHAP waterfall plots.
For each feature on the vertical axis, we represent its contribution (either negative, in blue, or positive, in red) to the final prediction for this particular patient. 

\subsection{Progression of Hospital Deployment}
We describe the progress of software use, from the pilot deployment with physician champions at the main hospital, and following the initial benefits revealed, expanded scales of utilization in all hospitals in the network.

\vspace{0.2mm}
\noindent \textit{Financial Benefits from Pilot Implementation at HA:} 
Before rolling out the tool to the other hospitals, the network's leadership and financial departments evaluated the potential benefits of the tool, based on the result of the pilot implementation at HA. 
They conduct a simple before-and-after analysis of the patient length of stay and estimate (see details in the \ref{ssec:a.pilot} Section of the Appendix) that the pilot implementation could have generated a \$711,348.44 increase in annual contribution margin, hence motivating further deployment. 

\vspace{0.2mm}
\noindent \textit{Large-scale Deployment in All Hospitals:}
Based on the positive feedback and benefits from the pilot implementation,  the hospital system decided to expand the program not just to physicians but also to medical service teams, case management, and nursing unit leadership. They deployed the tool to more floor units and included more physician champions in the study. 
The network progressively rolled out the software in more units of HA, as well as other hospitals, with a focus on medicine and cardiology (some of such units also contain surgical service), because such units cover most of the discharges and would benefit most from using the predictions. 
Section~\ref{ssec:a.rollout} of the Appendix provides details on the progressive deployment, including information on 15 units where our tool is fully deployed (with start dates between July 11, 2022 to January 15, 2023) and 12 units where the incorporation has not been completed (as of April 15, 2023).

\subsection{Empirical Effect on LOS Reduction} \label{ssec:empirical}
We analyze the impact of using our tool on patient length of stay. 
We consider two groups of units from Table~\ref{tab:unit_deployment} in the Appendix: a treatment group of 15 units that fully used our tool in January 15--April 15, 2023, and a control group of 12 units that had not yet fully incorporated the tool as of April 15, 2023. 
We collect discharge data for patients whose exit status is neither expired nor hospice since 2021 and compute the LOS (in number of days with decimal points) of each patient as the difference between admission order time and discharge time.
We exclude patients whose recorded admission time is after their recorded discharge time.
The outcome of interest is the average LOS of patients discharged from each unit group. 
This convention (i.e., assigning patients to their discharge unit only) is aligned with the hospitals' performance evaluation and with the fact that our tool mostly helps physicians anticipate and plan the discharge process. 

To estimate the effect of our tool, we  
use a Difference-in-Differences (DiD) technique~\citep{abadie2005semiparametric,bertrand2004how} and compare
the average change in LOS among patients in the treatment group to that of patients in the control group over time. 
We control for similar patient population fixed effects between the two groups, as they cover units of the same level of care (general level) and specialty (medicine and cardiology). We also control for the time non-stationarity effect within the year on LOS by focusing on the same January 15--April 15 period over the past three years.

\begin{figure}[tbp]
     \centering
     \includegraphics[width=\textwidth]{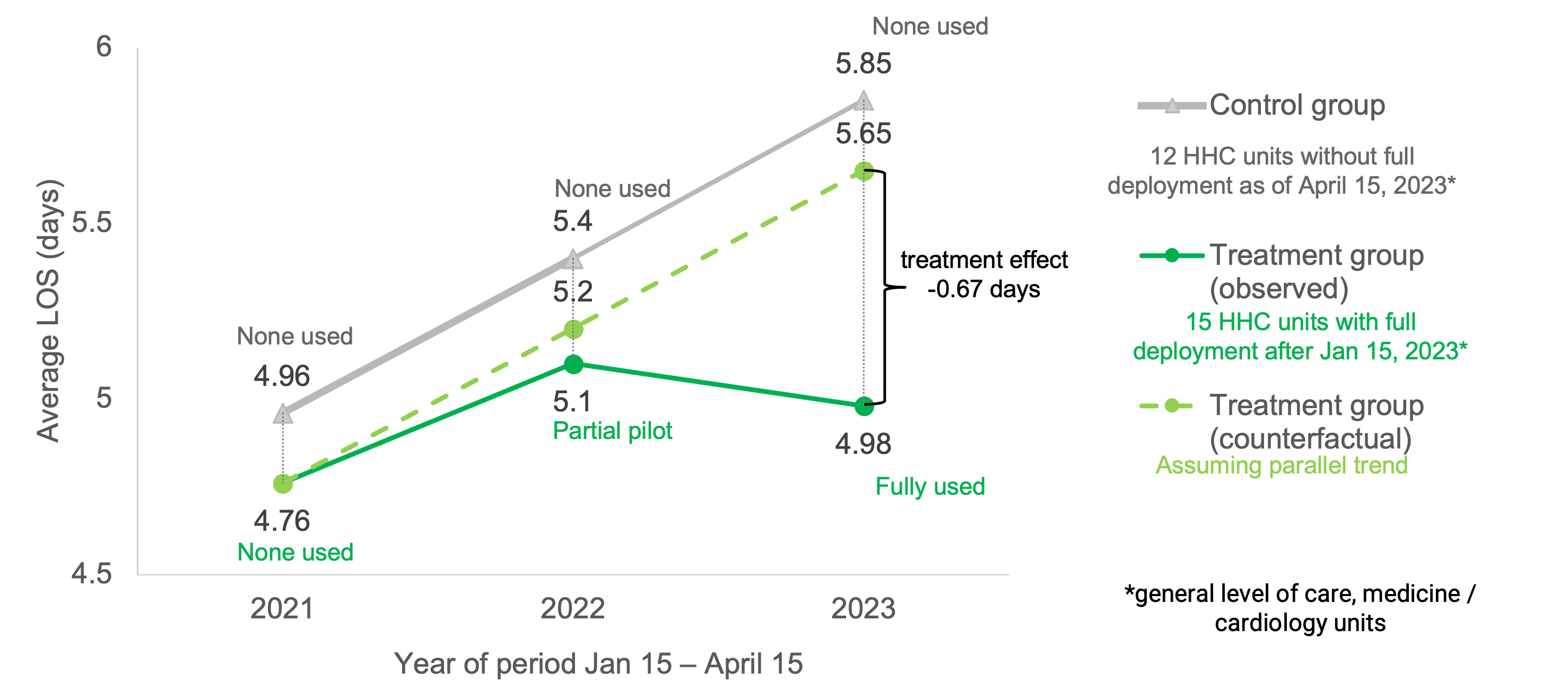}
    \caption{Empirical Analysis for Treatment Effect on Length of Stay.}
    \label{fig:los_did}
\end{figure}

We present the results in Figure~\ref{fig:los_did}, which shows the average LOS over the three-month period in 2021, 2022 and 2023. 
The control group did not use the tool throughout, whereas the treatment group had varying degrees of tool usage during the three periods. 
The hospital system reports no tool usage from January 15--April 15, 2021, partial tool usage in 5 units due to the pilot program from January 15--April 15, 2022, and full deployment of the tool in all 15 treatment units from January 15--April 15, 2023. 
We assume that the difference in LOS over time would have been the same between the two groups if the tool had not been used throughout the periods, based on the parallel trend assumption. 
We use this assumption to impute the counterfactual average LOS for the treatment group, had there been no treatment (in light green dashes).

The LOS of the control group showed a steady (approximately linear) increase, rising from 4.96 to 5.4 and eventually reaching 5.85 over three periods. Between 2021 and 2022, the treatment group's LOS increased from 4.76 to 5.1, which was in line with the parallel trend but slightly lower by 0.1 days, potentially due to the pilot's partial treatment effect.
After full deployment, the treatment group's LOS dropped to 4.98 in 2023, while the control group's LOS continued to rise from 4.96 to 5.85. The difference between the parallel counterfactual and actual treatment group's LOS resulted in an estimated benefit of reducing the average patient length of stay by 0.67 days. This benefit was more significant than the estimation from the pilot evaluation at HA due to several reasons, such as the tool's increased utilization in more units and hospitals and its deeper integration into the daily clinical workflow. 
Our analysis incorporated control variables such as time period of the year, level of care, and specialty of units, and it was multi-center, which was absent from the simple before-and-after comparison for the pilot. 

However, our analysis holds several assumptions and limitations. For instance, this is a descriptive analysis that assumed a parallel trend of LOS between the two equivalent patient populations, controlling only basic time and unit variables. 
Moreover, even though a small portion of physicians in the control units had access and used the predictions, we deemed these units as the control group, which could lead to a more conservative estimation of the effect. Finally, heavy-tailed LOS distributions could be significantly influenced by patient outliers with overly long LOS. 

\subsection{Projected Financial Benefits}
Table~\ref{tab:los_cm} outlines our projections for the estimated financial benefits on the network as a result of the LOS reduction. By reducing the average LOS by 0.67 days among patients in the 15 treatment units (49,424 patients annually), we can save 33,114.08 patient days. 
We consider two alternative scenarios, if 1) empty beds are not backfilled, and 2) empty beds are backfilled with new patients. 

Smaller hospitals with lower utilization rates would not backfill empty beds (resulted from the LOS reduction) with additional patients. Under a no backfill scenario, at an average direct cost for a medical/surgical inpatient of \$1,661 per patient day, the average annual savings from reducing the LOS by 0.67 days for patients in these units are estimated to be \$55,002,486.88. 

On the other hand, in larger hospitals like the main hospital who usually has a wait for their beds, the beds would be immediately backfilled. 
Under this alternative scenario, the benefits are transferred from cost savings into revenue increase by accepting additional patients. 
Assuming that all the beds would be backfilled, we estimate that this would make room for an additional 6,649.42 patients per year, with an average CM of \$10,796 per patient. This leads to a projected total annual CM increase of \$71,787,030.36 in these units. 

In practice, the hospitals are projected to obtain annual financial benefits in the range of \$55--\$72 million, with some beds backfilled and others otherwise.

\begin{table}[tbp]
\caption{Projection of Financial Benefits from LOS Reduction.}
\label{tab:los_cm}
\centering
\setlength\extrarowheight{-10pt}
\begin{tabular}{cc}
\toprule
\textbf{Deployment coverage:}           & 15 treatment units   \\ 
LOS reduction (days)          & 0.67 (estimated)   \\
Patients per year             & 49,424                      \\
Patient days saved            & 33,114.08                    \\
Average new LOS               & 4.98                            \\
\midrule
\textbf{Benefits if no beds backfilled:} & \\
Cost saving per patient & \$1112.87 \\ 
Total cost saving & \$55,002,486.88\\
\midrule
\textbf{Benefits if all beds backfilled:} & \\
Additional patients           & 6,649.42                     \\
Average CM per patient        & \$10,796           \\ 
Total CM increase             & \$71,787,030.36        \\ 
\end{tabular}
\end{table}

\subsection{Limitations and Future Work}
Although we tried to incorporate as many relevant variables as possible, some information considered predictive by the medical team is still  unavailable and unaccounted for by the current models. This includes medically-related variables (e.g., Clinical Institute Withdrawal Assessment for Alcohol scale score) as well as social and operational issues that are usually recorded in nursing and case management notes (e.g., palliative discussion, pending authentication to care facilities).
The team is working on obtaining these additional features and notes, where the latter would require supplemental Institutional Review Board protocol, text de-identification, and privacy management.

Like many medical institutions, the hospital network uses a commercial EMR system, from which we extract data using our own data pipeline (from Figure~\ref{fig:daily_pipeline}). Due to the technical complexity of this integration, data is refreshed once a day only and with an 8-hour delay. Physicians, however, identified that some relevant information might become available between the extraction time (midnight) and the time the data extracts become available (8 a.m.), information that the model currently cannot take into account. 
A more frequent update with shorter delays is an area of potential improvement, especially for ICU predictions.
As of today, despite extensive efforts, we have not been able to lift the technical bottlenecks in the hospital data system that generate these long delays.
We are currently working on running the models directly within the hospital EMR system (instead of conducting data extracts and running the models on a dedicated AWS server), which would allow for near-to-real-time data updates.

In future work, we will continue expanding the deployment of the tool and evaluating its impact on LOS and other patient/hospital outputs, as more data is collected over longer time periods. 
Followed by the connection between higher discharge predictions and lower readmission risks, another direction is to develop and integrate patient-level readmission risk predictions in production to enhance patient safety in practice. 

\section{Concluding Remarks} \label{sec:conclusion}
As part of a collaboration between a large hospital network, a academic research group, and a consultancy company, we develop a system of machine learning models predicting short-term discharge and ICU risk, as well as end-of-stay mortality and discharge disposition.
After three years of iterative development and validation, the final models aggregate a variety of clinical and nonclinical data about the daily status of each patient. All models achieve state-of-the-art predictive accuracy and are well calibrated. 

In particular, the 48-hour discharge predictions generated by the analytics program provide 6.3\%-29.6\% higher distinguishing power (measured by AUC) than the ones currently obtained from the healthcare providers directly, and the combination of the two (human and algorithmic) predictions can achieve 12.1\%-29.4\% higher precision or identify 10\%-28.7\% more discharge opportunities. 
Furthermore, we observe a negative relationship between the predicted discharge probabilities and 30-day and 7-day readmission risks, unintentionally capturing discharge safety. 
Patients discharged with green alerts have statistically significantly lower readmission rates compared with those without green alerts, suggesting another potential use of discharge predictions to enhance patient safety. 

In addition to being accurate, these models have been deployed in seven hospitals and are being used by hundreds of doctors, nurses, and case managers. 
The close and long-term partnership with key stakeholders at the hospital system enabled a broad adoption of the tool by medical providers and a deep integration within their daily workflow. 
As a result, the network experienced first-hand benefits to shorten the length of stay, decrease the cost of care, facilitate the education of patients and families regarding discharge, enhance patient safety, and improve the overall patient experience. 
Empirically, we observe a reduction in average patient length of stay by 0.67 days and project an annual contribution margin increase of \$55-\$72 million. 
The successful deployment in multiple centers of a representative U.S. hospital network also demonstrates the significant potential to scale the framework to other healthcare systems, in the U.S. and around the world.

\section*{Author Contributions}
L.N. led the efforts of model development, planning and performing experiments, writing code, analyzing results, and writing the manuscript, and contributed to data processing and model implementation. 
K.V.C. planned and performed experiments, wrote code, analyzed results, and wrote the manuscript. 
J.P. contributed to the research, experiment design, and writing of the manuscript. 
A.H-S. processed data, wrote code, maintained production, and contributed to model development and software implementation. 
D.K. supervised the medical team, directed deployment at the hospitals, contributed to model development and evaluation, and edited the manuscript.
M.B-L., A.C., M.K, and P.H. provided medical feedback and contributed to model improvement and implementation at the hospital. 
B.S. supervised the medical team and hospital deployment. 
D.B. directed the overall project, from concept and research to implementation, and edited the manuscript.

\ACKNOWLEDGMENT{
The authors would like to thank the Hartford HealthCare team for their help with implementation, feedback, and discussions.
In particular, we are grateful to Frank Damiano, Qun Yu, and Michelle Schneider for their support in data extraction.
The authors thank Kyle Maulden, Yi Wang, and Yiwen Zhang for providing support on computational experiments to our work. 
Finally, we acknowledge the MIT SuperCloud and Lincoln Laboratory Supercomputing Center for providing computing resources and technical consultation.
}

%
%
%
\newpage
\begin{APPENDICES}
\section{Hospitals in the Network} \label{sec:a.hhc}
The main hospital of the network, Hospital A
(HA), is one of the largest teaching hospitals in New England, in collaboration with a medical university.
HA is a tertiary hospital, recognized to be high performing in a variety of procedures, conditions, and specialties.
Hospital B (HB) is an acute-care community teaching hospital and a trauma center in the east region, with an outstanding specialty in stroke.
Hospital C (HC) is a general acute care community hospital serving a  healthcare resource in the northwest region.
Hospital D (HD) is a central region community teaching hospital providing comprehensive inpatient and outpatient services in a variety of specialties and participating in residency programs with a medical school.
Hospital E (HE) is another community hospital serving the central region with a wide collection of services.
Hospital F (HF) is a tertiary community teaching hospital and mission-driven Catholic hospital that provides care with special attention to the most vulnerable and poor patients.
Hospital G (HG) is a community hospital in the east region and does not provide an ICU level of care.
Descriptive information and statistics about the seven hospitals are provided in  Table~\ref{tab:hhc_stats}. 

\begin{table}[tbp]
\centering  \footnotesize
\caption{Descriptive Statistics about the 7 hospitals in 2021.}
\label{tab:hhc_stats}
\setlength\extrarowheight{-10pt}
\begin{tabular}{ccccccc} 
Hospital & \# Beds & \# Units & \# Services & Highest Level & patient days & Operating Revenue \\ \midrule
HA   & 867 & 47   & 48  & Intensive Care    & 261,954  & \$2 billion   \\ 
HB   & 233 & 13   & 38  & Critical Care     & 52,328   & \$449.9 million   \\ 
HC   & 122 & 11   & 18  & Intensive Care    & 27,912   & \$175 million \\ 
HD & 446 & 20   & 34  & Critical Care     & 76,325   & \$552.8 million   \\ 
HE  & 156 & 12   & 31  & Intensive Care    & 39,972   & \$385.4 million   \\ 
HF   & 520 & 26   & 43  & Intensive Care    & 85,322   & \$466.5 million   \\ 
HG   & 130 & 5    & 14  & Step Down     & 11,545   & \$127.5 million   \\ 
\end{tabular}
\end{table}

\section{Data and Feature Processing} \label{sec:a.featureprocessing}
In this Appendix section, we elaborate on the data extract and feature processing, and in particular, changes and efforts taken to expand models from one hospital to seven hospitals.

\vspace{0.2mm}  \noindent \textit{Data Extracts:}
We build 10 data extracts summarized in Table~\ref{tab:data_extract}.
\begin{table}[tbp]
\small
\caption{Summary of Data Extracts.}
\label{tab:data_extract}
\setlength\extrarowheight{-10pt}
\begin{tabular}{clll}
{Extract \# }     & {Extract description}       & {Granularity} & {Example columns}      \\ \midrule
1     & Admission, discharge, transfer events & Event    & Hospital, department destination     \\ 
2    & Admission, discharge, transfer orders & Order    & Service, order type      \\ 
3     & Lab results with normal ranges    & Patient day  & Platelet count with normal range  \\ 
4  & Clinical measurements     & Patient day  & Blood pressure, respiratory rate \\ 
5 & Preparation for discharge     & Patient day  & Discharge time, future surgery date \\ 
6  & DNR status   & Patient day  & DNR          \\
7    & Time invariant patient information    & Patient  & Age, discharge disposition  \\ 
8    & Summary statistics of notes   & Patient day  & Diagnosis, number of notes written   \\ 
9   & Surgery       & Surgery case & Procedure name, start time, end time     \\ 
10   & Summary statistics of orders  & Patient day  & Number of orders, pending labs \\ 
\end{tabular}
\end{table}

\vspace{0.2mm}  \noindent \textit{Full List of Curated Features:}
We create various features divided into six groups.
\begin{enumerate}
    \item[1)] Current conditions:
    \begin{itemize}
    \item Information extracted from admission, discharge, transfer orders, and events (e.g., department, service, whether in ICU).
    \item Current statuses such as DNR and Nothing by Mouth (NPO, meaning inability to eat or drink).
    \item Others such as dialysis and the oxygen (O2) device.
    \end{itemize}
    \item[2)] Lab results (e.g., albumin, white blood cell count):
    \begin{itemize}
    \item Latest measurements.
    \item Delta variables (difference of the current day’s lab result from the previous day's).
    \item Normal range indicators (1 if the lab value is within the normal range, 2 if outside the normal range, and 0 if the lab value is missing).
    \item Distance between the lab value and its normal range, i.e., min(value-lower bound, 0)+max(value-upper bound, 0). 
    \item Number of abnormal lab results in the past 24 hours.
    \end{itemize}
    \item[3)] Clinical measurements (e.g., temperature, respiratory rate, heart rate):
    \begin{itemize}
    \item Latest measurement.
    \item Highest measurement in the past 24 hours.
    \item Average value in the past 24 hours.
    \item Critical indicators (whether or not the value is critical with respect to the critical range provided by doctors).
    \end{itemize}
    \item[4)] Time series summary statistics of operational variables:
    \begin{itemize}
    \item Number of days in the ICU and in hospitalization since admission.
    \item Number of days until the next scheduled surgery, the number of days since the last surgery, and total time spent in surgery since admission.
    \item Pending results (whether or not MRI/CT/ECHO/etc. is pending, and the number of labs pending).
    \item Numbers of notes, orders, and medications in the last 24 hours and since admission.
   \item Number of attending physicians in the last 24 hours.
   \end{itemize}
    \item[5)] Patient information prior to current admission:
    \begin{itemize}
    \item Age at admission and patient type.
    \item Number of days since the previous admission, and LOS of the previous admission.
    \end{itemize}
    \item[6)] Auxiliary operational variables which are not patient-specific:
    \begin{itemize}
    \item Current date-related variables such as day of the week and weekend indicator.
    \item Ward-related statistics such as census, utilization, and daily discharges.
    \item Hospital-related statistics (e.g. number of hospital admissions in the past 24 hours).
    \end{itemize}
\end{enumerate}
\vspace{0.2mm}

\noindent \textit{Missing Feature Imputation:}
Like most clinical data, our data extracts contain a large portion of missing data.
We adopt different imputation techniques to deal with different types of missing features.
First, with the hospital's help, we leverage information about how the data are collected, recorded, and computed to develop a rule that deterministically imputes a subset of variables.
For some binary variables in group 1),  a missing value indicates that the patient does not have this current condition. For example, DNR is missing when the patient did not sign a DNR, NPO is missing when the patient does not have an NPO constraint, and IV is missing when the patient is not on IV. Thus we fill the categorical values accordingly for such binary variables. 
For some multi-class categorical variables in group 1), missing values are deemed as a separate category indicating that it is missing. For example, missing O2 device, and diagnosis are imputed as ``NA Value'' to indicate the patient has no O2 device and has no diagnosis.
For the range columns of lab results in group 2) (e.g. normal range for bilirubin), we first extract the non-missing record for each admission event and backfill on all dates for this admission event. Here we assume that the normal range column has unique input for each admission event. If all records for the admission event are missing, we fill the entries with the most frequent category of this range column.
For clinical measurements in group 3), we capture cases when a value is not measured, which could reflect additional information, by imputing the missing value with a special value such as -1.
For most summary statistics of operational variables in group 4, we fill the count with 0 if there is no record found. For example, if there is no pending lab and no medications in the record, then the count of pending labs and the number of medications are computed as 0.
For some other counting variables in groups 4) and 5), imputing with 0 would confuse with a count of 0 due to their meanings. For example, the number of days until the next scheduled surgery is 0 if the future surgery date is the current day. Thus we impute a missing future surgery date, which means the patient does not have surgery scheduled, with the value -1 instead of 0. 
For the number of days since the last surgery and since the previous admission, no record of the last surgery and admission is approximated with a long time ago by imputing with 9999.
After our clinical rule-based imputation step, other variables such as laboratory results and vital devices also have a high missing percentage.
We drop columns with more than 50\% missing values in the dataset and then impute missing values of the remaining features using the OptImpute~\citep{bertsimas2017predictive}, an imputation method based on optimization and k nearest neighbors~\citep{peterson2009k}.
We note that for features in group 3), we impute the original lab results and clinical measurements before computing the delta and distance to the normal range.
\vspace{0.2mm}

\noindent \textit{String Parsing:}
Several columns in the data extracts require string parsing to obtain numerical values.
The normal range features are stored in the format of  ``lower bound - upper bound'', ``$>$upper bound'', or ``$<$lower bound'' as strings. 
Hence, we first split these strings into two float formatted numbers as upper bound and lower bound features. 
Several variables such as the last RASS score and last pain score take string formats containing both the numerical score and an explanation of the score, such as ``0 $\to$ alert and calm'', ``-4 $\to$ deep sedation'', ``10 $\to$ hurts worst'', etc.
We parse such string columns to extract only the numerical score as a continuous variable.
\vspace{0.2mm}

\paragraph{Categorical Variable Encoding.}
The feature space contains several categorical variables and requires encoding before passing into the imputation and modeling process. Some features must be encoded in a specific way, for example, the feature DNR is encoded as 1 if the entry is ``DNR (DO NOT RESUSCITATE)'', and 0 otherwise. For other categorical variables, such as current department, current service, and oxygen device, we use a label encoder to encode each column separately. 
\vspace{0.2mm}

\paragraph{Differences between hospitals.}
Since the electronic medical records of the seven hospitals are unified in one system, most variables have consistent forms across hospitals and thus did not require additional processing to scale from one hospital to others.
However, each hospital has some different conventions and some variables required specific processing for each hospital.
For example, hospitals have different ways of naming departments and levels of care.
Departments of intensive care level are named ICU in HA, HC, and HE, named CCU in HB, and Critical Care in HD, HG does not have an ICU, and HF have specific names for ICU units.
These differences were identified by inspection of the data and consultation with the hospital network.
Thus we modified the way to compute ICU-related variables such as whether or not a patient is in the ICU and the number of days in the ICU, as well as to compute the ICU-related prediction targets for each hospital differently.
The hospital is set to be a parameter to address the differences in the feature processing part.
Another major difference is that multi-class categorical variables can take very different sets of values from one hospital to another.
This could be due to differences in both patient populations and in ways of recording data by different staff members.
In previous models for HA only, some multi-class categorical variables, such as service, diagnosis, insurance, etc., were manually encoded based on their critical levels where similar values of variables are grouped together. 
For example, O2 devices are converted into numerical values ranging from 1 (most critical devices) to 7 (room air or no device).
Such encoding was done with discussions with doctors at HA based on their clinical knowledge.
However, since the seven hospitals have different categories, manual categorization would need to be done for each hospital individually.
To make the process more efficient and scalable, we decided to replace such manual encoding with label encoding, where we have one encoder for each categorical variable for each hospital. 
The encoders are saved for each hospital for consistent use in production.
Similarly, we train, save, and apply separate OptImpute imputers for each hospital.
\vspace{0.2mm}

\section{Supplementary Results} \label{sec:a.resuts}
In this Appendix section, we present additional results of the data and experiments.

\subsection{Data Summary Statistics} \label{ssec:a.stats}
We report here the summary statistics of the data after the process of inclusion, exclusion, and splits from the Machine Learning Modeling section.
For each hospital, the number of patients, admissions, and patient days in the union of training, validation, and testing sets are summarized in Table~\ref{tab:data_size}.
In total, we use data from 180,682 patients, 280,593 admissions, and 1,375,215 patient days.
The data sizes vary across hospitals, where HA, the hospital with largest data size, has over 33 times of patient days than HG, the hospital with the smallest data size.
After filtering per target, the numbers of data points, i.e., patient days of the remaining training, validation, and testing sets combined are shown in Table~\ref{tab:target_size}. 
Mortality and discharge-related targets keep the majority of the data points, whereas ICU targets have fewer data points largely due to the split of patients in versus not in ICU, especially leaving ICU predictions have a small portion of data because a very small portion of patients are in ICU.
\vspace{-1em}

\begin{table}[h]
\centering  \footnotesize
\setlength\extrarowheight{-10pt}
\caption{Summary of Data Size.}
\label{tab:data_size}
\begin{tabular}{llllllll}
Hospital & HA& HB  & HC   & HD    & HE    & HF & HG \\ \midrule
\# Patients & 105,184 & 15,493  & 7,956  & 20,011  & 15,576 & 11,624 & 4,838  \\ 
\# Admissions & 171,072  & 23,354  & 12,822  & 29,490  & 21,612 & 15,319 & 6,924  \\ 
\# patient days & 879,357 & 106,662 & 52,931 & 139,542 & 90,924 & 79,615 & 26,184 \\ 
\end{tabular}
\end{table}

\begin{table}[tbp]
\centering \footnotesize
\setlength\extrarowheight{-10pt}
\caption{Number of Data Points (patient days) for Each Prediction.}
\label{tab:target_size}
\begin{tabular}{llllllll}
Hospital Prediction  & HA  & HB  & HC  & HD    & HE    & HF & HG      \\ \midrule
Mortality    & 865,954  & 104,552 & 51,542 & 134,684 & 88,999 & 76,618 & 25,823      \\ 
Discharge Disposition & 865,954 & 104,552 & 51,542 & 134,684 & 88,999 & 76,618 & 25,823      \\ 
Discharge 24hr   & 869,468 & 105,451 & 52,021  & 135,592 & 89,474 & 77,178 & 25,937      \\ 
Discharge 48hr   & 868,563 & 105,225 & 51,851  & 135,297 & 89,320 & 77,032 & 25,888      \\ 
Enter ICU 24 hr   & 592,264 & 75,585  & 34,378 & 92,700  & 63,876 & 59,701 & \multirow{4}{*}{No ICU} \\ 
Leave ICU 24 hr  & 115,997  & 7,808   & 5,792  & 17,499  & 5,526  & 4,590  &     \\ 
Enter ICU 48 hr   & 454,803 & 56,679  & 24,166 & 69,097  & 47,491 & 47,569 &     \\ 
Leave ICU 48 hr   & 109,443 & 7,312   & 5,336   & 16,332  & 5,079  & 4,360  &     \\ 
\end{tabular}
\end{table}

We also report the proportions observed of each prediction task outcome in the testing set.
For mortality and discharge disposition, since the target outcome is the outcome at the end of the stay, we compute the proportions of patients for each of the three discharge disposition classes for each hospital in Table~\ref{tab:target_prop_admissions}.
The proportions are similar between hospitals, except that HB and HG have a higher proportion of patients discharged to the home without service category compared to the other hospitals.
For the discharge and ICU next 24-hr and 48-hr targets, since the outcome depends on the date, we compute the proportions of patient days that have a positive outcome for each prediction task in each hospital in Table~\ref{tab:target_prop}.
Compared with the other 6 hospitals, HG has a significantly higher proportion of positive discharge 24-hr / 48-hr outcomes, as HG tends to treat less critical patients without the presence of ICU. 
Entering ICU predictions have highly imbalanced classes, as less than 3\% of patients enter the ICU in the next 24 and 48 hours.
The proportion of patients leaving ICU ranges from 77.41\% to 88.92\% for 24 hours and from 64.19\% to 80.60\% between the six hospitals, likely due to the different composition of the patient population and congestion level in each hospital.

\begin{table}[tbp]
\centering \footnotesize
\caption{Proportion of Patient Admissions for Each Discharge Disposition in Testing Set.}
\label{tab:target_prop_admissions}
\setlength\extrarowheight{-10pt}
\begin{tabular}{llllllll}
Outcome Class   & HA   & HB  & HC  & HD    & HE & HF  & HG  \\ \midrule
Expired / hospice    & 5.63\% & 6.08\%  & 7.47\%   & 7.17\%  & 6.31\%  & 7.41\%  & 5.12\%  \\ 
Home without service & 50.04\% & 60.44\% & 45.26\% & 50.16\% & 52.51\% & 52.08\% & 61.97\% \\ 
With service     & 44.33\% & 33.48\% & 47.27\% & 42.67\% & 41.18\% & 40.51\% & 32.91\% \\ 
\end{tabular}
\end{table}

\begin{table}[tbp]
\centering \footnotesize
\setlength\extrarowheight{-10pt}
\caption{Proportion of patient days with Positive Target Outcomes in Testing Set.}
\label{tab:target_prop}
\begin{tabular}{p{4cm}lllllll}
Prediction Target & HA  & HB  & HC  &  HD    & HE & HF  & HG      \\ \midrule
Discharge 24hr    & 17.94\% & 19.81\% & 20.56\% & 18.61\% & 21.08\% & 17.44\% & 25.75\%     \\ 
Discharge 48hr    & 32.95\% & 36.54\% & 37.60\%  & 34.49\% & 38.05\% & 31.58\% & 46.12\%     \\ 
Enter ICU 24 hr   & 1.62\%  & 0.93\%  & 1.35\%  & 1.53\%  & 0.69\%  & 0.90\%  & \multirow{4}{*}{No ICU} \\ 
Leave ICU 24 hr   & 81.04\%  & 83.34\% & 88.92\% & 84.06\% & 85.69\% & 77.41\% &     \\ 
Enter ICU 48 hr  & 2.78\%  & 1.62\%  & 2.52\%   & 2.48\%  & 1.23\%  & 1.46\%  &     \\ 
Leave ICU 48 hr  & 69.89\%  & 73.16\% & 80.60\% & 74.35\% & 76.40\% & 64.19\% &     \\ 
\end{tabular}
\end{table}

\FloatBarrier
\subsection{Assessment of Model Calibration} \label{ssec:a.calibration}
We evaluate the proper calibration of all our models on the second half of the testing set, which was not used during calibration. This evaluation is performed using calibration curves, which compare the probabilities predicted by the calibrated model vs the empirical probabilities of the data. Following \cite{degroot1983comparison, niculescu2005predicting}, we generate these curves by bucketing the probabilities predicted by the calibrated model into 10 uniform bins. Within each bin/bucket, we compare the empirical probability and the average predicted probability (or classification score).
We then plot the resulting points (we only include points for bins with at least 10 observations, to reduce noise). As observed from sample calibration curves in Figure~\ref{fig:calibration_icu}, the points mostly fall near the diagonals, indicating that the final models are well calibrated. As shown in Table \ref{tab:calibration_sse}, all models are indeed well calibrated; in particular, the mean squared errors (averages of the squares of the differences between the predicted probabilities and the empirical probabilities across the nonempty bins) for all models are below $0.01$.
\begin{figure}[tbp]
 \centering
\begin{subfigure}[b]{0.49\textwidth}
 \centering
 \includegraphics[width=0.85\textwidth]{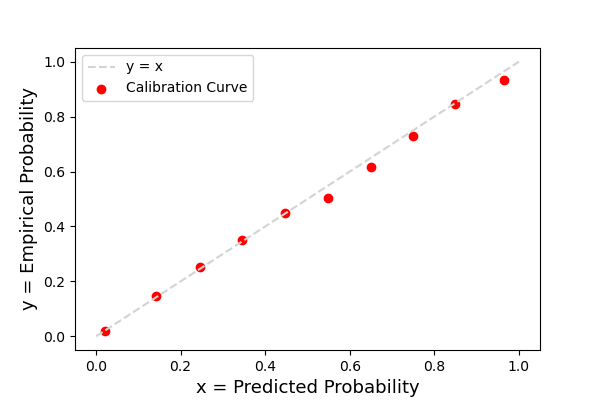}
 \caption{HA Mortality.}
 \label{subfig:hhmortality_c}
 \end{subfigure}
 \hfill
 \begin{subfigure}[b]{0.49\textwidth}
 \centering
 \includegraphics[width=0.85\textwidth]{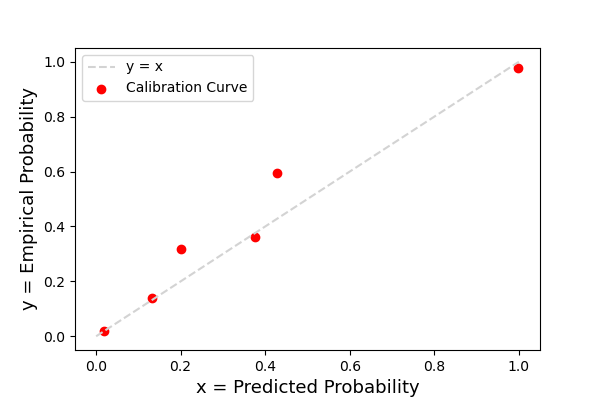}
  \caption{HG Mortality.}
 \label{subfig:whmortality_c}
 \end{subfigure}
 \begin{subfigure}[b]{0.49\textwidth}
 \centering
 \includegraphics[width=0.85\textwidth]{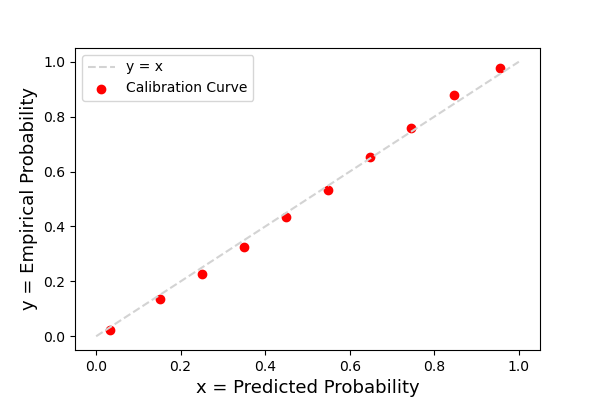}
 \caption{HA Discharge 48hr.}
 \label{subfig:hh_dis_c}
 \end{subfigure}
 \hfill
 \begin{subfigure}[b]{0.49\textwidth}
 \centering
 \includegraphics[width=0.85\textwidth]{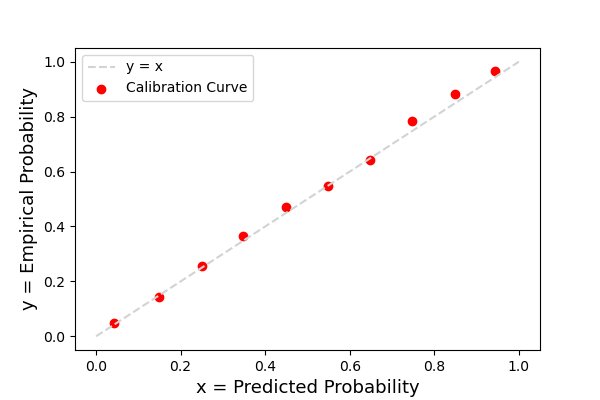}
  \caption{HF Discharge 48hr.}
 \label{subfig:sv_dis_c}
 \end{subfigure}
 %
\begin{subfigure}[b]{0.49\textwidth}
 \centering
 \includegraphics[width=0.85\textwidth]{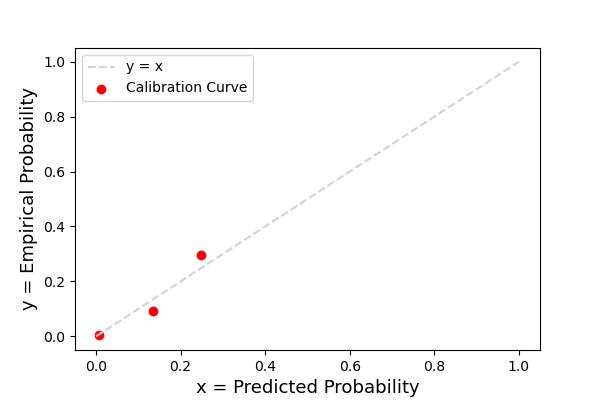}
 \caption{HB Enter ICU 24hr.}
 \end{subfigure}
 \hfill
 \begin{subfigure}[b]{0.49\textwidth}
 \centering
 \includegraphics[width=0.85\textwidth]{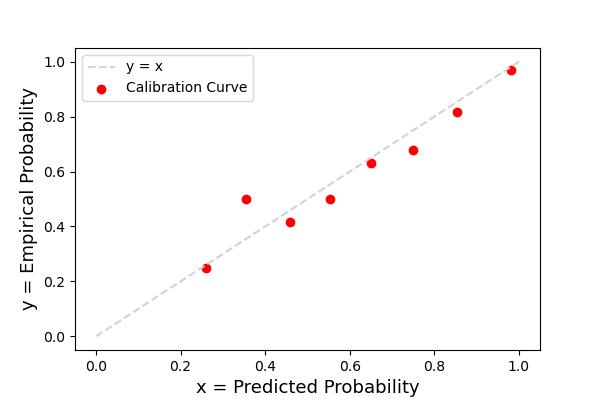}
  \caption{HE Leave ICU 24hr.}
 \end{subfigure}
 \begin{subfigure}[b]{0.49\textwidth}
 \centering
 \includegraphics[width=0.85\textwidth]{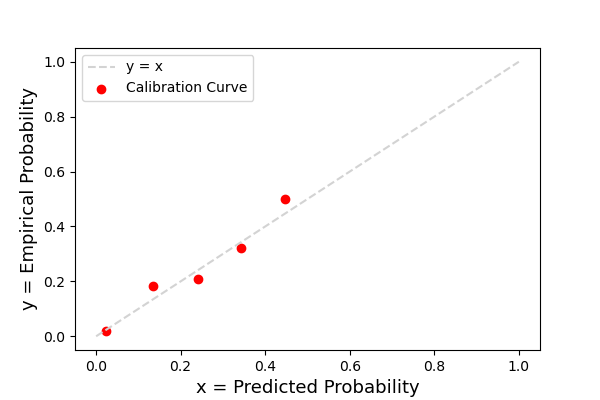}
 \caption{HD Enter ICU 48hr.}
 \end{subfigure}
 \hfill
 \begin{subfigure}[b]{0.49\textwidth}
 \centering
 \includegraphics[width=0.85\textwidth]{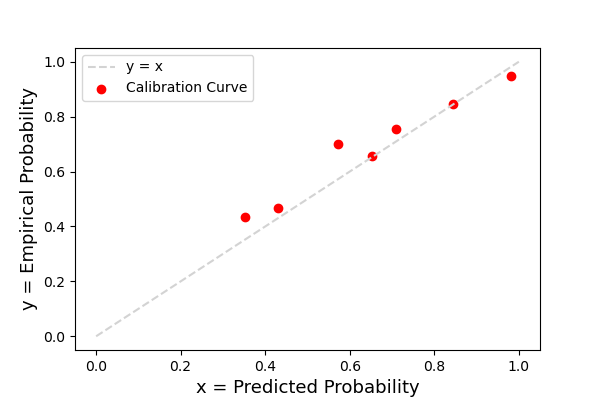}
  \caption{HC Leave ICU 48hr.}
 \end{subfigure}
\caption{Calibration Curves for Prediction Models.}
\label{fig:calibration_icu}
\end{figure}

\begin{table}[tbp]
    \centering
    \setlength\extrarowheight{-10pt}
     \caption{Mean Squared Errors for the Calibration Curve of Each Model with Nonempty Bins.}
    \label{tab:calibration_sse}
    \begin{tabular}{p{4cm}lllllll}
{Prediction Target} &      HA & HB    & HC      & HD  & HE   & HF    & HG  \\
\midrule
      Mortality & 6.94e-05 & 7.71e-04 & 8.88e-04 & 6.94e-04 & 1.96e-03 & 2.33e-03 & 1.17e-03 \\
Discharge 24 hr & 1.46e-04& 2.06e-04 & 1.25e-03  & 1.92e-04 & 4.86e-04 & 1.73e-04 & 4.77e-04 \\
Discharge 48 hr & 2.69e-04& 7.29e-04 & 6.96e-04  & 4.77e-04 & 1.07e-03 & 2.34e-04 & 1.99e-03 \\
Enter ICU 24 hr & 1.67e-04& 7.27e-05 & 9.65e-05  & 2.50e-05 & 7.26e-07 & 2.30e-06 &    \multirow{4}{*}{No ICU} \\
Leave ICU 24 hr & 5.50e-04& 2.87e-03 & 6.44e-03  & 2.61e-03 & 7.17e-04 & 4.87e-03 &     \\
Enter ICU 48 hr & 1.77e-04& 5.60e-05 & 2.39e-04  & 1.00e-04 & 3.12e-05 & 3.40e-05 & \\
Leave ICU 48 hr & 9.28e-04& 4.27e-03 & 2.09e-03  & 7.33e-04 & 7.63e-03 & 6.50e-03 &      \\
\end{tabular}
\end{table}

\FloatBarrier

\subsection{Accuracy of the Color-Coded Alert System} \label{ssec:a.alert}
In this section, we evaluate the alerts (green and red) for each hospital. 

First, we report the accuracy, precision, and recall for our previous green ($t_{24} = t_{48}=0.5$), new green ($t_{24} = 0.25, t_{48}=0.4$), and red ($t = 0.2, t_{\delta}=0.1$) alerts at all seven hospitals in Table~\ref{tab:sel_prec_rec}.
Among the hospitals, previous green alerts have 0.696-0.782 accuracy, 0.645-0.698 precision, and 0.546-0.684 recall; new green alerts have 0.687-0.768 accuracy, 0.588-0.629 precision, and 0.701-0.8 recall; red alerts have 0.885-0.925 accuracy, 0.477-0.55 precision, and 0.471-0.715 recall.

\begin{table}[h]
\centering  \footnotesize
\setlength\extrarowheight{-10pt}
\caption{Precision and Recall under Selected Thresholds for Alerts.}
\label{tab:sel_prec_rec}
\begin{tabular}{ccccccccc}
Alert       & Hospital   & HA & HB    & HC      & HD  & HE   & HF    & HG    \\\midrule
\multirow{3}{*}{Previous green}  & Accuracy & 0.782 & 0.757 & 0.739  & 0.771 & 0.759 & 0.78  & 0.696 \\
        & Precision & 0.698 & 0.672 & 0.645 & 0.682 & 0.684 & 0.692 & 0.68  \\
        & Recall    & 0.598 & 0.653 & 0.679 & 0.628 & 0.684 & 0.546 & 0.646 \\
        \midrule
\multirow{3}{*}{New green} & Accuracy & 0.767  & 0.734 & 0.714 & 0.751 & 0.739 & 0.768 & 0.687 \\
        & Precision & 0.621 & 0.604 & 0.588  & 0.611 & 0.623 & 0.617 & 0.629 \\
        & Recall     & 0.746 & 0.786 & 0.8   & 0.764 & 0.796 & 0.701 & 0.78  \\
        \midrule
\multirow{3}{*}{Red}    & Accuracy & 0.899 & 0.901 & 0.895  & 0.881 & 0.896 & 0.886 & 0.925 \\
        & Precision & 0.477  & 0.55  & 0.574 & 0.492 & 0.528 & 0.505 & 0.53  \\
        & Recall  & 0.705  & 0.668 & 0.553  & 0.691 & 0.715 & 0.663 & 0.471\\ 
\end{tabular}
\end{table}

\subsection{SHAP Summary Plots} \label{ssec:a.shap}
In Figure~\ref{fig:shap_icu}, we present SHAP summary plots for discharge and ICU predictions for six different hospitals.
We analyze Figure~\ref{subfig:hh_dis} and Figure~\ref{subfig:sv_dis} for discharge 48-hr models.
Some major clinical discharge barriers are identified, such as intensive care, fall risk score, NPO, O2 device, O2 concentration, and future surgery date, which are used as top variables in both models' and doctors' predictions.
In addition to the clinical variables, the models use a variety of operational variables that also have significant contributions to the predictions.
A lot of these variables are time-series variables such as counting the number of abnormal blood results, lab orders, ICD codes, days in hospitalization, notes, etc. in the past 24 hours or since admission.
Others are not patient-specific; for example, the day of the week and daily discharges from the ward often affect discharge probabilities as well.
Compared with models, doctors focus mainly on the clinical aspects of the patients.
The models learn that discharge depends on a combination of clinical and operational characteristics of the patients as well as affected by the hospital's operational status, which is the case in practice as well.
Moreover, by comparing the plots between different hospitals in Figure~\ref{fig:shap}, we observe that models for different hospitals find common important variables.

For entering ICU, the importance of the future surgery date and the service aligns with known hospital protocols; for example, patients with scheduled cardiology surgery typically go to the ICU after surgery.
Many of the significant features identified are similar to those for mortality prediction as entering ICU is strongly correlated with increased mortality risk.
For leaving ICU, top features include some oxygen-related variables, such as O2 device, O2 concentration, and SPO2, as well as other clinical variables such as RASS measurement and inverse flow.
\begin{figure}[tbp]
 \centering
  \begin{subfigure}[b]{0.49\textwidth}
     \centering
     \includegraphics[width=\textwidth,height=5cm]{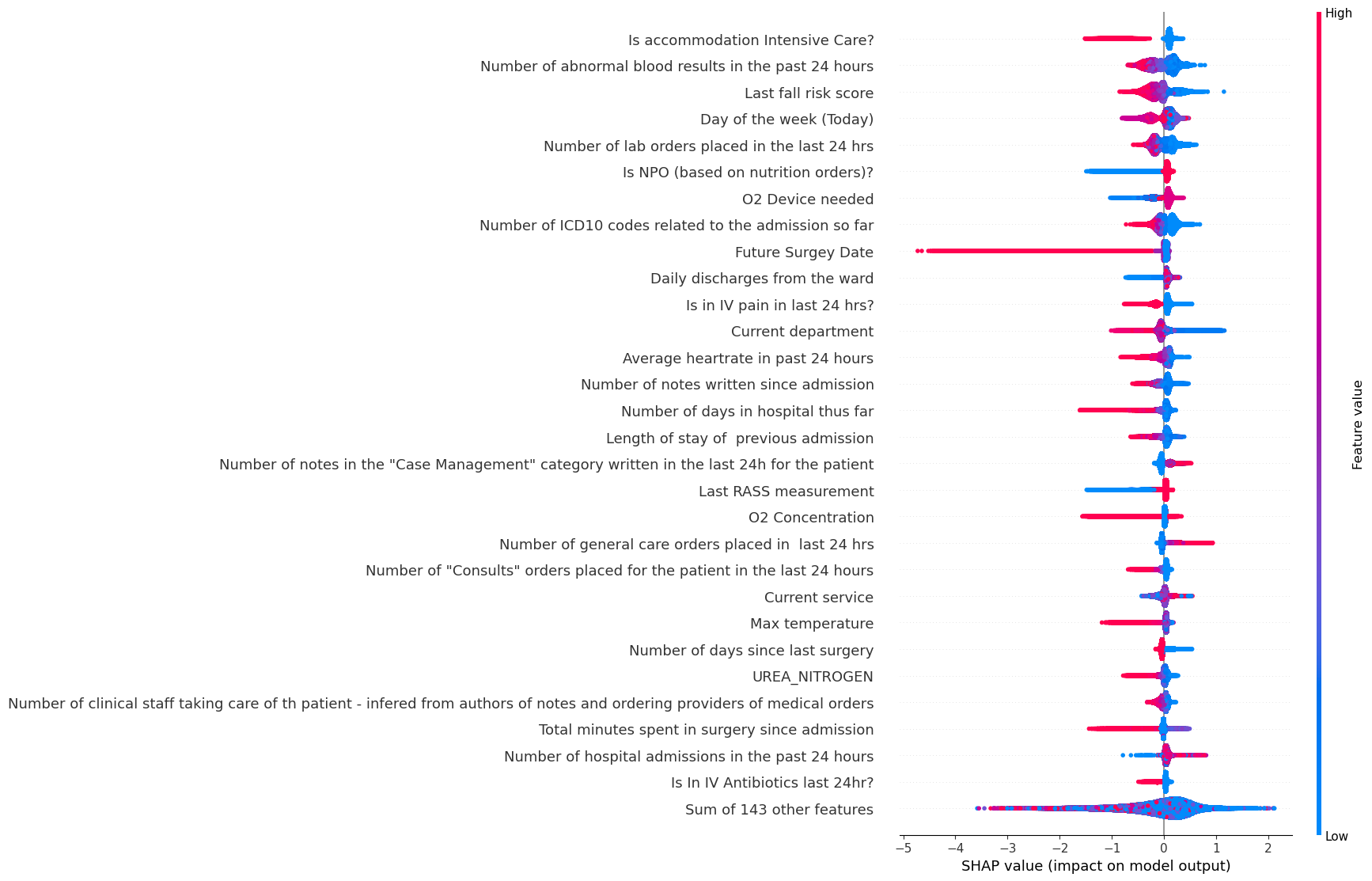}
     \caption{HA Discharge 48hr.}
     \label{subfig:hh_dis}
 \end{subfigure}
 \hfill
 \begin{subfigure}[b]{0.49\textwidth}
     \centering
     \includegraphics[width=\textwidth,height=5cm]{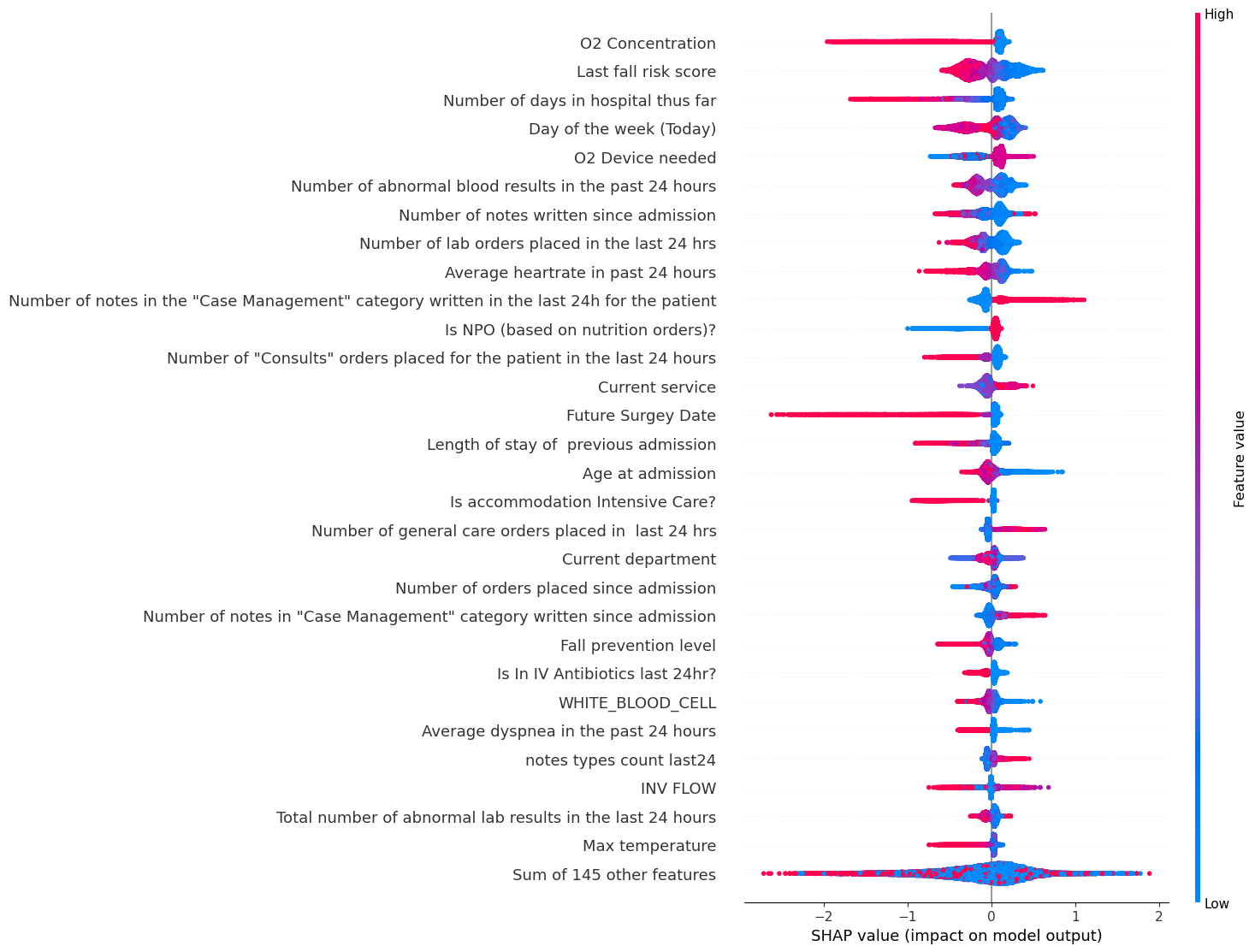}
  \caption{HF Discharge 48hr.}
     \label{subfig:sv_dis}
 \end{subfigure}
    \begin{subfigure}[b]{0.49\textwidth}
     \centering
     \includegraphics[width=\textwidth,height=5cm]{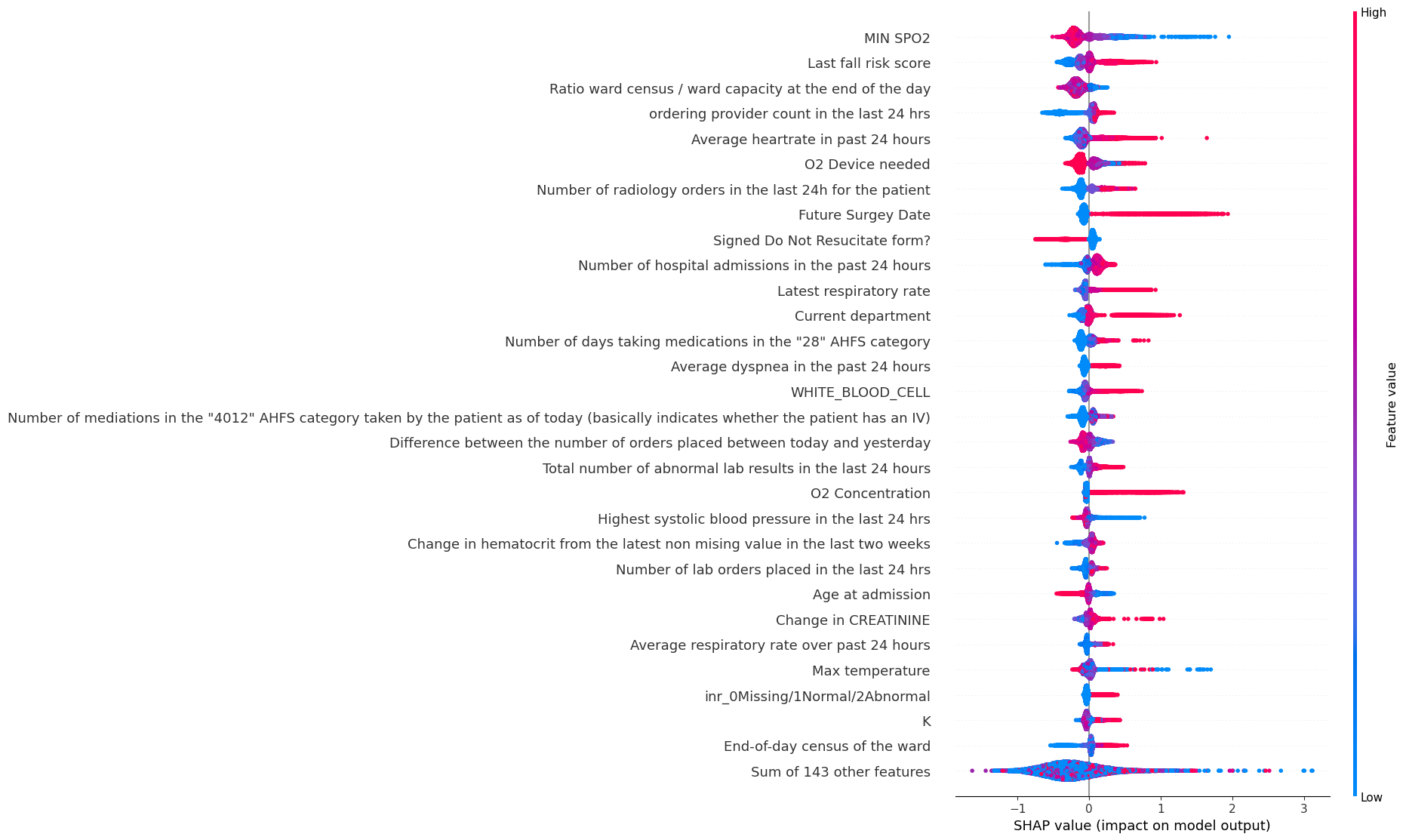}
     \caption{HB Enter ICU 24hr.}
 \end{subfigure}
 \hfill
 \begin{subfigure}[b]{0.49\textwidth}
     \centering
     \includegraphics[width=\textwidth,height=5cm]{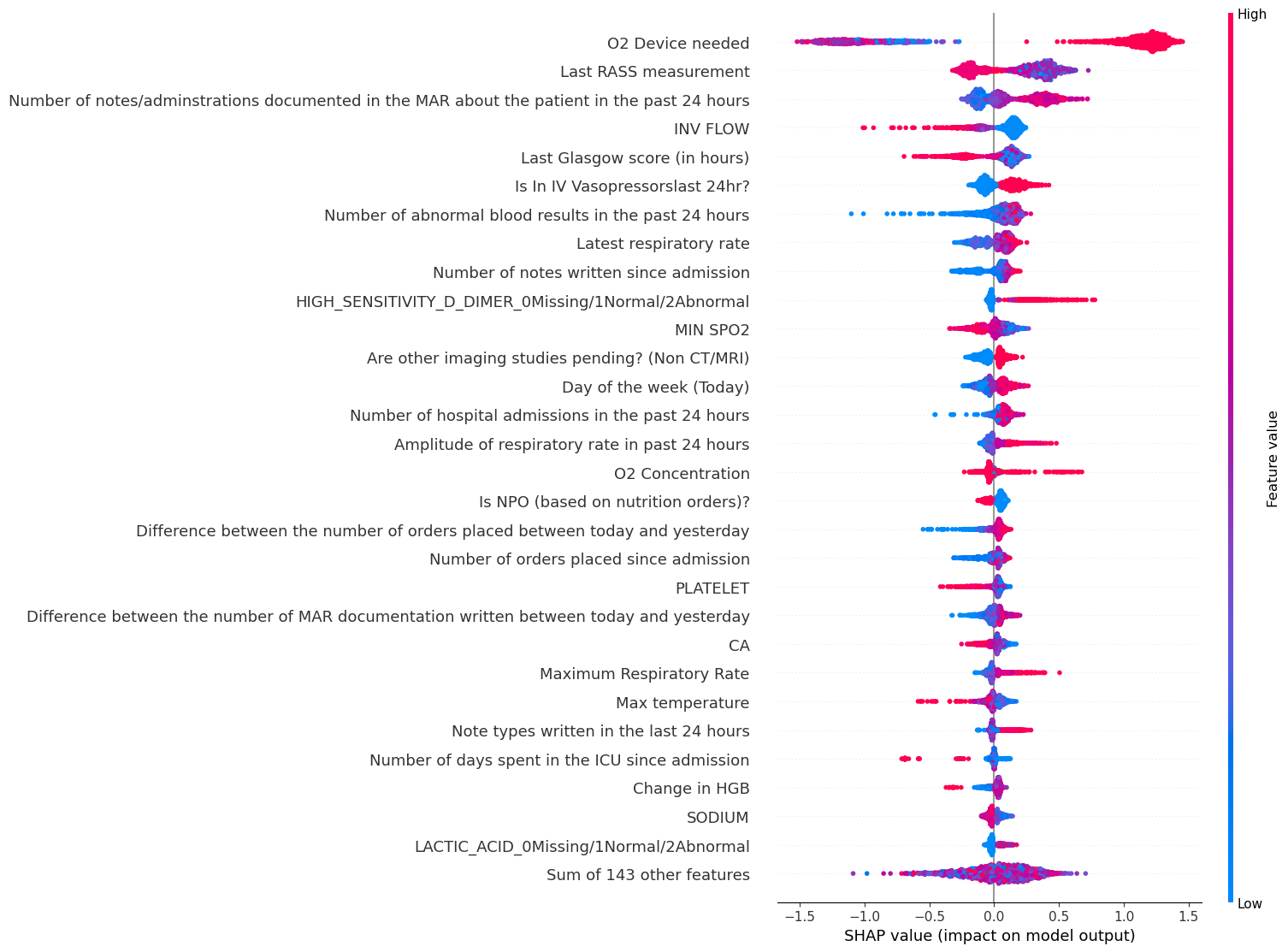}
  \caption{HE Leave ICU 24hr.}
 \end{subfigure}
 \begin{subfigure}[b]{0.49\textwidth}
     \centering
     \includegraphics[width=\textwidth,height=5cm]{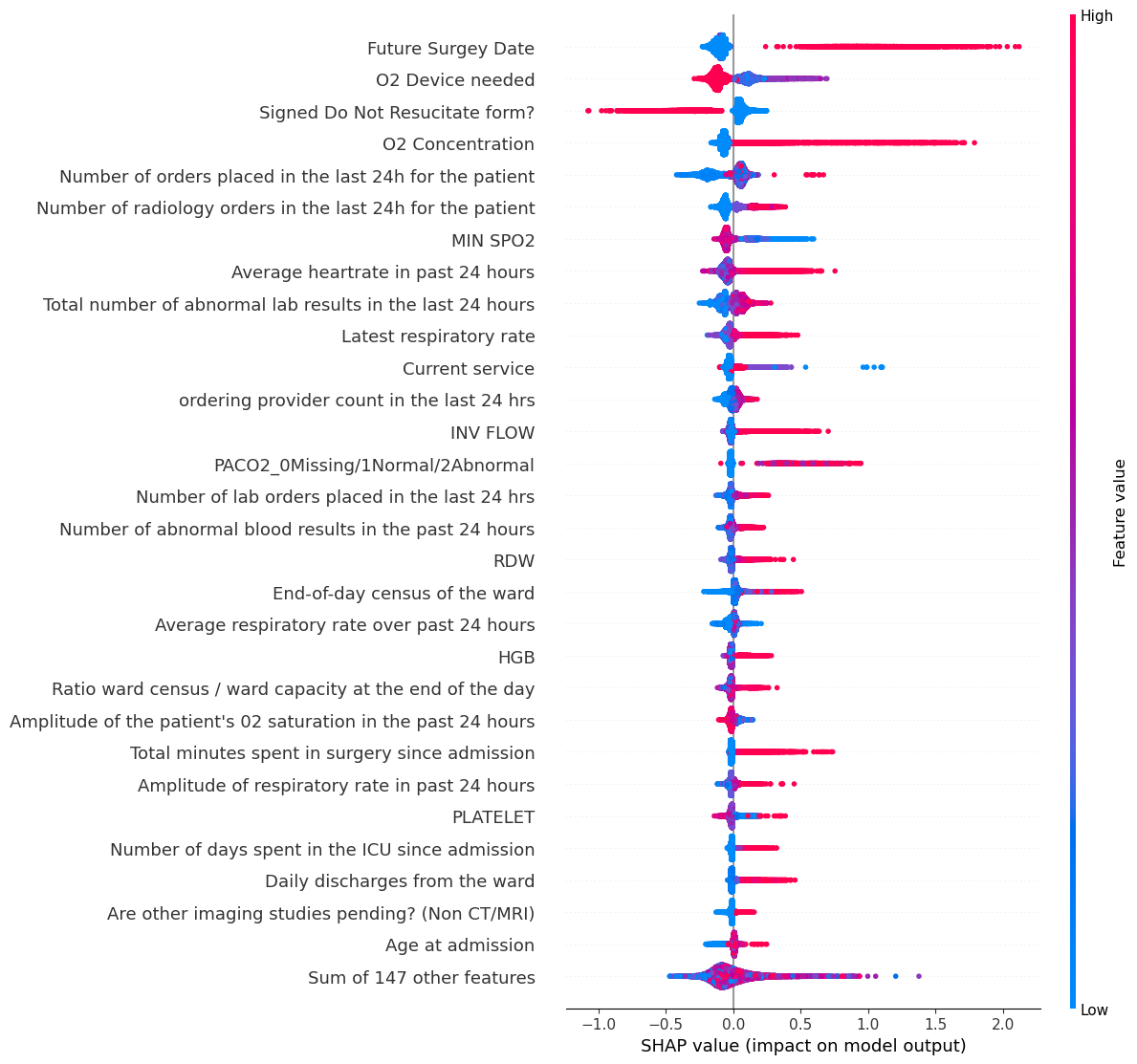}
     \caption{HD Enter ICU 48hr.}
 \end{subfigure}
 \hfill
 \begin{subfigure}[b]{0.49\textwidth}
     \centering
     \includegraphics[width=\textwidth,height=5cm]{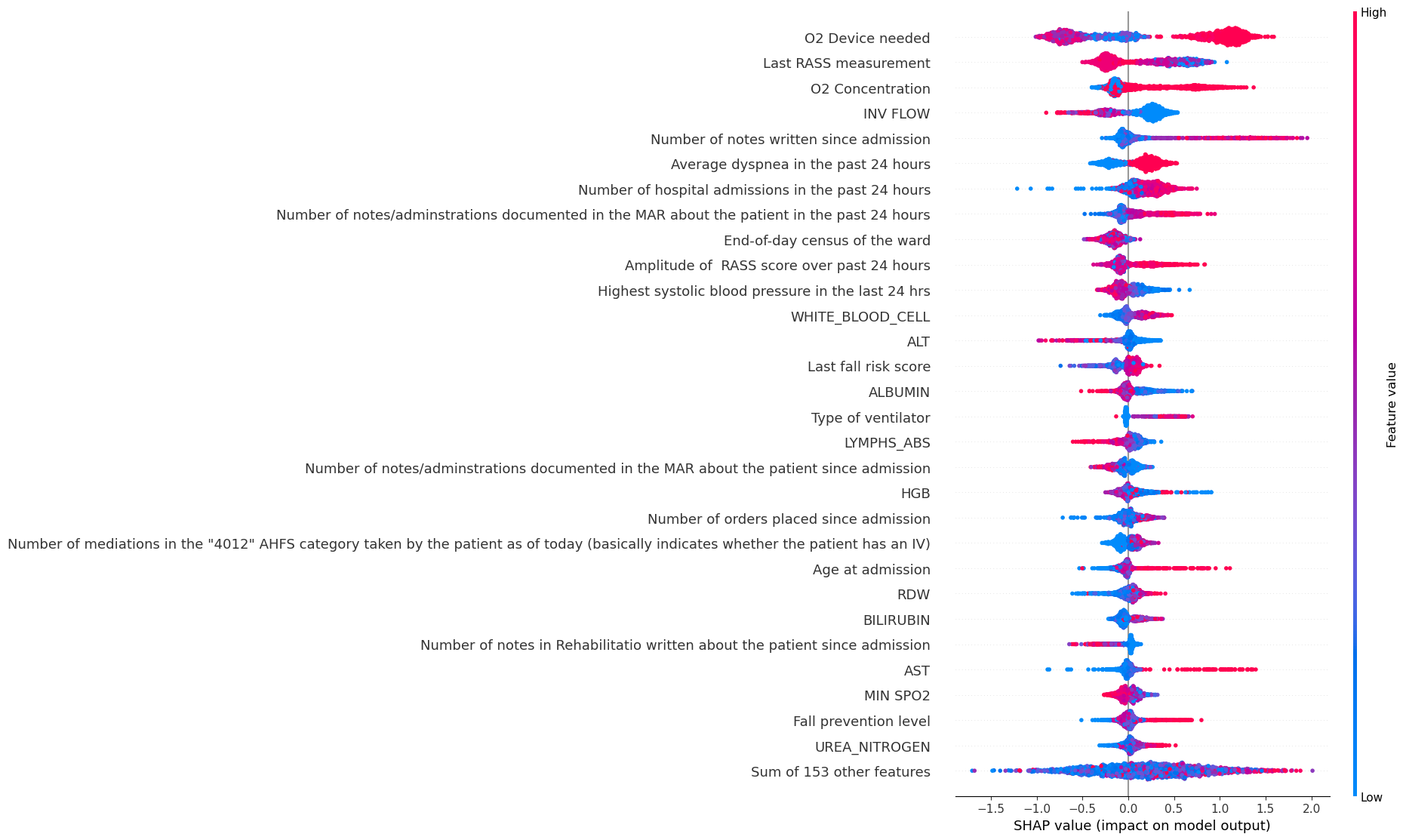}
  \caption{HC Leave ICU 48hr.}
 \end{subfigure}
    \caption{SHAP Summary Plots for Discharge and ICU Predictions.}
    \label{fig:shap_icu}
\end{figure}

\FloatBarrier
\section{Supplementary Information on Implementation and Impact} \label{sec:a.impact}
In this Appendix section, we present additional information and results of the implementation and impact.

\subsection{Estimation of the Financial Benefits from Pilot} \label{ssec:a.pilot}
To evaluate the benefits of the pilot program, the main hospital comparse the lengths of stay of patients in Q4 2020 (277 such patients) and Q4 2021 (351 patients), for those patients whose attending discharge physician is one of the four physician champions who had access to the tool in the second half of 2021. 
Compared with Q4 2020, they observe a reduction in average LOS by 0.35 days (from 5.84 to 5.49) in Q4 2021.
Given HA's high utilization rate, they assume that any additional bed made available thanks to a reduction in LOS will be occupied. 
Had the tool been available in Q4 2020, these 277 patients would have been discharged 0.35 days earlier, resulting in savings of 96.95 patient days for the quarter, extrapolated to 387.80 patient days for one year. 
At an average LOS of 5.84 in Q4 2020, the additional 387.80 patient days available translate into 65.89 additional patients.
At an average contribution margin of \$10,796 per patient, HA estimated an annual contribution margin increase of \$711,348.44 as a result of the pilot implementation. 

However, we acknowledge that this initial evaluation has several limitations. The pilot implementation only involves a limited number of physicians/units/patients, in a single hospital, and over a limited period of time. The magnitude of the benefits is likely to be different once the tool is more deeply integrated into the workflows of all medical staff, and deployed to other units and hospitals. 
Moreover, in their before-after analysis, the hospital compares LOS in 2020 Q4  and 2021 Q4 without controlling for potential confounding factors that could explain differences between the two periods (e.g., COVID-19 level).

\subsection{Staggered Roll-Out of Our Tool} \label{ssec:a.rollout}
Table~\ref{tab:unit_deployment} presents information and deployment progress of all units that satisfy the level and specialty of care criterion in the seven hospitals. 
During the second half of 2022, some units in four hospitals (HA--HD) fully incorporated the software in their daily clinical decision process since the start dates indicated in the table.  
Two more hospitals (HE and HG) began adopting the tool in several units on January 15, 2023 as the next phase of implementation. 
By April 15, 2023, 15 units across six hospitals had fully integrated the predictions in their review process, where unit leads review the predictions with the provider team daily and adjust decisions accordingly. 
Smaller hospitals with standard procedures of progression rounds with unit leads (HB, HC, and HG) deployed the tool in most of their eligible units, while other hospitals (HA, HD, HE) are still in the process of rolling out more units. 
Since HF does not have regular progression rounds where the medical team conducts a structured daily patient review process, a streamlined integration of our tool in their progression rounds is ongoing work. 
As of April 15, 2023, 12 other units (with an NA Start Date) had not officially integrated the daily process deeply, but individual physicians still have the option to access and use the predictions.

\begin{table}[tbp]
\centering 
\small 
\setlength\extrarowheight{-10pt}
\caption{Unit Deployment Progress Information.}
\label{tab:unit_deployment}
\begin{tabular}{ccccc}
Hospital & Unit     & Start Date & Specialty   & Capacity \\ \midrule
HA   & HA CONKLIN 2 & NA     & Medicine/Oncology  & 27      \\
HA   & HA CONKLIN 4 & 9/13/22    & Medicine    & 25      \\
HA   & HA CONKLIN 5 & 7/11/22    & Medicine    & 47      \\
HA   & HA BLISS 7 EAST  & 8/23/22    & Medicine    & 17      \\
HA   & HA BLISS 10 EAST & NA     & Cardiology  & 14      \\
HA   & HA CENTER 10 & NA     & Cardiology  & 26      \\
HA   & HA CENTER 12 & 7/11/22    & Medicine    & 26      \\
HA   & HA NORTH 10  & NA     & Cardiology  & 27      \\
HA   & HA NORTH 12  & 7/11/22    & Medicine    & 20  \\
\midrule
HB   & HB A3 MEDSURG    & 8/23/22    & Medicine/Surgical    & 30      \\
HB   & HB E4 Cardiology & 8/23/22    & Cardiology  & 28      \\
HC   & HC FOURTH FLOOR  & 8/23/22    & Medicine/Surgical    & 28      \\
HC   & HC FIFTH FLOOR   & 8/23/22    & Medicine/Surgical    & 29      \\
HD & HD EAST 2  & NA     & Medicine/Observation & 12      \\
HD & HD WEST 2  & NA     & Medicine     & 15      \\
HD & HD NORTH 3 & 1/15/23   & Medicine     & 24      \\
HD & HD NORTH 4 & 10/22/22   & Medicine/Cardiology  & 28      \\
HD & HD NORTH 5 & 8/23/22    & Medicine/Stroke  & 30      \\
\midrule
HE & HE PAVILION D & NA &	Medicine & 28 \\
HE & HE PAVILION E & 1/15/23	& Medicine	& 28\\
HF & HF 6 NORTH & NA & Cardiology & 20 \\
HF & HF 6 SOUTH & NA & Cardiology & 20 \\
HF & HF 9 NORTH & NA & Medicine & 22\\
HF & HF 10 NORTH & NA & Medicine & 29 \\
HG 	& HG 4 SHEA EAST &	1/15/23	& Medicine/Surgical	& 30\\
HG &	HG 4 SHEA NORTH	& 1/15/23	& Medicine/Surgical	& 12\\
HG &	HG GREER	& NA &			Medicine/Surgical &	23\\ 
\end{tabular}
\end{table}
\FloatBarrier
\end{APPENDICES}

\bibliographystyle{informs2014} 
\bibliography{bibliography.bib} 

\begin{thebibliography}{38}
\providecommand{\natexlab}[1]{#1}
\providecommand{\url}[1]{\texttt{#1}}
\providecommand{\urlprefix}{URL }

\bibitem[{Abadie(2005)}]{abadie2005semiparametric}
Abadie A (2005) Semiparametric difference-in-differences estimators. \emph{The
  Review of Economic Studies} 72(1):1--19.

\bibitem[{Arik \protect\BIBand{} Pfister(2021)}]{arik2021tabnet}
Arik S{\"O}, Pfister T (2021) {TabNet}: Attentive interpretable tabular
  learning. \emph{Proceedings of the AAAI Conference on Artificial
  Intelligence} 35(8):6679--6687.

\bibitem[{Awad et~al.(2017{\natexlab{a}})Awad, Bader-El-Den, \protect\BIBand{}
  McNicholas}]{awad2017patient}
Awad A, Bader-El-Den M, McNicholas J (2017{\natexlab{a}}) Patient length of
  stay and mortality prediction: a survey. \emph{Health Services Management
  Research} 30(2):105--120.

\bibitem[{Awad et~al.(2017{\natexlab{b}})Awad, Bader-El-Den, McNicholas,
  \protect\BIBand{} Briggs}]{awad2017early}
Awad A, Bader-El-Den M, McNicholas J, Briggs J (2017{\natexlab{b}}) Early
  hospital mortality prediction of intensive care unit patients using an
  ensemble learning approach. \emph{International Journal of Medical
  Informatics} 108:185--195.

\bibitem[{Bardak \protect\BIBand{} Tan(2021)}]{bardak2021improving}
Bardak B, Tan M (2021) Improving clinical outcome predictions using convolution
  over medical entities with multimodal learning. \emph{Artificial Intelligence
  in Medicine} 117:102112.

\bibitem[{Bertrand et~al.(2004)Bertrand, Duflo, \protect\BIBand{}
  Mullainathan}]{bertrand2004how}
Bertrand M, Duflo E, Mullainathan S (2004) How much should we trust
  differences-in-differences estimates? \emph{The Quarterly Journal of
  Economics} 119(1):249--275.

\bibitem[{Bertsimas \protect\BIBand{} Dunn(2017)}]{bertsimas2017optimal}
Bertsimas D, Dunn J (2017) Optimal classification trees. \emph{Machine
  Learning} 106:1039--1082.

\bibitem[{Bertsimas et~al.(2021{\natexlab{a}})Bertsimas, Pauphilet, Stevens,
  \protect\BIBand{} Tandon}]{bertsimas2021predicting}
Bertsimas D, Pauphilet J, Stevens J, Tandon M (2021{\natexlab{a}}) Predicting
  inpatient flow at a major hospital using interpretable analytics.
  \emph{Manufacturing \& Service Operations Management} 24(6):2809--2824.

\bibitem[{Bertsimas et~al.(2021{\natexlab{b}})Bertsimas, Pauphilet,
  \protect\BIBand{} Van~Parys}]{bertsimas2017sparse}
Bertsimas D, Pauphilet J, Van~Parys B (2021{\natexlab{b}}) Sparse
  classification: a scalable discrete optimization perspective. \emph{Machine
  Learning} 110(11):3177--3209.

\bibitem[{Bertsimas et~al.(2018)Bertsimas, Pawlowski, \protect\BIBand{}
  Zhuo}]{bertsimas2017predictive}
Bertsimas D, Pawlowski C, Zhuo YD (2018) From predictive methods to missing
  data imputation: an optimization approach. \emph{Journal of Machine Learning
  Research} 18(196):1--39.

\bibitem[{Chen \protect\BIBand{} Guestrin(2016)}]{chen2015xgboost}
Chen T, Guestrin C (2016) {XGBoost}: A scalable tree boosting system.
  \emph{Proceedings of the 22nd {ACM} SIGKDD International Conference on
  Knowledge Discovery and Data Mining}, 785–794 (Association for Computing
  Machinery).

\bibitem[{Christ et~al.(2018)Christ, Braun, Neuffer, \protect\BIBand{}
  Kempa-Liehr}]{christ2018time}
Christ M, Braun N, Neuffer J, Kempa-Liehr AW (2018) Time series feature
  extraction on basis of scalable hypothesis tests (tsfresh--a python package).
  \emph{Neurocomputing} 307:72--77.

\bibitem[{Covino et~al.(2020)Covino, Sandroni, Santoro, Sabia, Simeoni, Bocci,
  Ojetti, Candelli, Antonelli, Gasbarrini, \protect\BIBand{}
  Franceschi}]{covino2020predicting}
Covino M, Sandroni C, Santoro M, Sabia L, Simeoni B, Bocci MG, Ojetti V,
  Candelli M, Antonelli M, Gasbarrini A, Franceschi F (2020) Predicting
  intensive care unit admission and death for {COVID-19} patients in the
  emergency department using early warning scores. \emph{Resuscitation}
  156:84--91.

\bibitem[{DeGroot \protect\BIBand{} Fienberg(1983)}]{degroot1983comparison}
DeGroot MH, Fienberg SE (1983) The comparison and evaluation of forecasters.
  \emph{Journal of the Royal Statistical Society: Series D (The Statistician)}
  32(1-2):12--22.

\bibitem[{{Interpretable AI, LLC}(2022)}]{InterpretableAI}
{Interpretable AI, LLC} (2022) Interpretable {AI} documentation.
  \urlprefix\url{https://www.interpretable.ai}.

\bibitem[{Jencks et~al.(2009)Jencks, Williams, \protect\BIBand{}
  Coleman}]{jencks2009rehospitalizations}
Jencks SF, Williams MV, Coleman EA (2009) Rehospitalizations among patients in
  the medicare fee-for-service program. \emph{New England Journal of Medicine}
  360(14):1418--1428.

\bibitem[{Jin et~al.(2018)Jin, Bahadori, Colak, Bhatia, Celikkaya, Bhakta,
  Senthivel, Khalilia, Navarro, Zhang, Doman, Ravi, Liger, \protect\BIBand{}
  Kass-hout}]{jin2018improving}
Jin M, Bahadori MT, Colak A, Bhatia P, Celikkaya B, Bhakta R, Senthivel S,
  Khalilia M, Navarro D, Zhang B, Doman T, Ravi A, Liger M, Kass-hout T (2018)
  Improving hospital mortality prediction with medical named entities and
  multimodal learning. \emph{arXiv preprint arXiv:1811.12276} .

\bibitem[{Ke et~al.(2017)Ke, Meng, Finley, Wang, Chen, Ma, Ye,
  \protect\BIBand{} Liu}]{ke2017lightgbm}
Ke G, Meng Q, Finley T, Wang T, Chen W, Ma W, Ye Q, Liu TY (2017) Lightgbm: A
  highly efficient gradient boosting decision tree. \emph{Advances in Neural
  Information Processing Systems} 30.

\bibitem[{Kim et~al.(2016)Kim, Chan, Olivares, \protect\BIBand{}
  Escobar}]{kim2016association}
Kim SH, Chan CW, Olivares M, Escobar GJ (2016) Association among {ICU}
  congestion, {ICU} admission decision, and patient outcomes. \emph{Critical
  Care Medicine} 44(10):1814--1821.

\bibitem[{Koekkoek et~al.(2011)Koekkoek, Bayley, Brown, \protect\BIBand{}
  Rustvold}]{koekkoek2011hospitalists}
Koekkoek D, Bayley KB, Brown A, Rustvold DL (2011) Hospitalists assess the
  causes of early hospital readmissions. \emph{Journal of Hospital Medicine}
  6(7):383--388.

\bibitem[{Long \protect\BIBand{} Mathews(2018)}]{long2018boarding}
Long EF, Mathews KS (2018) The boarding patient: effects of {ICU} and hospital
  occupancy surges on patient flow. \emph{Production and Operations Management}
  27(12):2122--2143.

\bibitem[{Mathews et~al.(2018)Mathews, Durst, Vargas-Torres, Olson, Mazumdar,
  \protect\BIBand{} Richardson}]{mathews2018effect}
Mathews K, Durst M, Vargas-Torres C, Olson A, Mazumdar M, Richardson L (2018)
  Effect of emergency department and {ICU} occupancy on admission decisions and
  outcomes for critically ill patients. \emph{Critical Care Medicine}
  46(5):720–727.

\bibitem[{Mees et~al.(2016)Mees, Klein, Yperzeele, Vanacker, \protect\BIBand{}
  Cras}]{mees2016predicting}
Mees M, Klein J, Yperzeele L, Vanacker P, Cras P (2016) Predicting discharge
  destination after stroke: a systematic review. \emph{Clinical Neurology and
  Neurosurgery} 142:15--21.

\bibitem[{Niculescu-Mizil \protect\BIBand{}
  Caruana(2005)}]{niculescu2005predicting}
Niculescu-Mizil A, Caruana RA (2005) Predicting good probabilities with
  supervised learning. \emph{Proceedings of the 22nd International Conference
  on Machine Learning}, 625--632.

\bibitem[{Peterson(2009)}]{peterson2009k}
Peterson LE (2009) K-nearest neighbor. \emph{Scholarpedia} 4(2):1883.

\bibitem[{Rajkomar et~al.(2018)Rajkomar, Oren, Chen, Dai, Hajaj, Hardt, Liu,
  Liu, Marcus, Sun, Sundberg, Yee, Zhang, Zhang, Flores, Duggan, Irvine, Le,
  Litsch, Mossin, Tansuwan, Wang, Wexler, Wilson, Ludwig, Volchenboum, Chou,
  Pearson, Madabushi, Shah, Butte, Howell, Cui, Corrado, \protect\BIBand{}
  Dean}]{rajkomar2018scalable}
Rajkomar A, Oren E, Chen K, Dai AM, Hajaj N, Hardt M, Liu PJ, Liu X, Marcus J,
  Sun M, Sundberg P, Yee H, Zhang K, Zhang Y, Flores G, Duggan GE, Irvine J, Le
  Q, Litsch K, Mossin A, Tansuwan J, Wang D, Wexler J, Wilson J, Ludwig D,
  Volchenboum SL, Chou K, Pearson M, Madabushi S, Shah NH, Butte AJ, Howell MD,
  Cui C, Corrado GS, Dean J (2018) Scalable and accurate deep learning with
  electronic health records. \emph{{npj} Digital Medicine} 1(18).

\bibitem[{Safavi et~al.(2019)Safavi, Khaniyev, Copenhaver, Seelen, Langle,
  Zanger, Daily, Levi, \protect\BIBand{} Dunn}]{safavi2019development}
Safavi KC, Khaniyev T, Copenhaver M, Seelen M, Langle ACZ, Zanger J, Daily B,
  Levi R, Dunn P (2019) Development and validation of a machine learning model
  to aid discharge processes for inpatient surgical care. \emph{JAMA Network
  Open} 2(12):e1917221--e1917221.

\bibitem[{Saleh et~al.(2020)Saleh, Makam, Halm, \protect\BIBand{}
  Nguyen}]{saleh2020can}
Saleh SN, Makam AN, Halm EA, Nguyen OK (2020) Can we predict early 7-day
  readmissions using a standard 30-day hospital readmission risk prediction
  model? \emph{BMC Medical Informatics and Decision Making} 20(227):1--7.

\bibitem[{{\c{S}}enol et~al.(2013){\c{S}}enol, Saylam, Kocaay,
  \protect\BIBand{} Tez}]{csenol2013red}
{\c{S}}enol K, Saylam B, Kocaay F, Tez M (2013) Red cell distribution width as
  a predictor of mortality in acute pancreatitis. \emph{The American Journal of
  Emergency Medicine} 31(4):687--689.

\bibitem[{Soni \protect\BIBand{} Gopalakrishnan(2021)}]{soni2021significance}
Soni M, Gopalakrishnan R (2021) Significance of {RDW} in predicting mortality
  in {COVID-19}—an analysis of 622 cases. \emph{International Journal of
  Laboratory Hematology} 43(4):O221–O223.

\bibitem[{{\v{S}}trumbelj \protect\BIBand{}
  Kononenko(2014)}]{vstrumbelj2014explaining}
{\v{S}}trumbelj E, Kononenko I (2014) Explaining prediction models and
  individual predictions with feature contributions. \emph{Knowledge and
  Information Systems} 41:647--665.

\bibitem[{Subudhi et~al.(2021)Subudhi, Verma, Patel, Hardin, Khandekar, Lee,
  McEvoy, Stylianopoulos, Munn, Dutta, \protect\BIBand{}
  Jain}]{subudhi2021comparing}
Subudhi S, Verma A, Patel AB, Hardin CC, Khandekar MJ, Lee H, McEvoy D,
  Stylianopoulos T, Munn LL, Dutta S, Jain RK (2021) Comparing machine learning
  algorithms for predicting {ICU} admission and mortality in {COVID-19}.
  \emph{{npj} Digital Medicine} 4(87).

\bibitem[{Wang et~al.(2018)Wang, Ma, Kao, Tsai, \protect\BIBand{}
  Chang}]{wang2018red}
Wang AY, Ma HP, Kao WF, Tsai SH, Chang CK (2018) Red blood cell distribution
  width is associated with mortality in elderly patients with sepsis. \emph{The
  American Journal of Emergency Medicine} 36(6):949--953.

\bibitem[{Wang et~al.(2022)Wang, Hussain, Birge, Schreiber, \protect\BIBand{}
  Adelman}]{wang2022high}
Wang K, Hussain W, Birge JR, Schreiber MD, Adelman D (2022) A high-fidelity
  model to predict length of stay in the neonatal intensive care unit.
  \emph{INFORMS Journal on Computing} 34(1):183--195.

\bibitem[{Zadrozny \protect\BIBand{} Elkan(2002)}]{zadrozny2002transforming}
Zadrozny B, Elkan C (2002) Transforming classifier scores into accurate
  multiclass probability estimates. \emph{Proceedings of the Eighth ACM SIGKDD
  International Conference on Knowledge Discovery and Data Mining}, 694--699.

\bibitem[{Zhao et~al.(2020)Zhao, Chen, Hou, Graham, Li, Richman, Thode, Singer,
  \protect\BIBand{} Duong}]{zhao2020prediction}
Zhao Z, Chen A, Hou W, Graham JM, Li H, Richman PS, Thode HC, Singer AJ, Duong
  TQ (2020) Prediction model and risk scores of {ICU} admission and mortality
  in {COVID-19}. \emph{PlOS One} 15(7):e0236618.

\bibitem[{Zhou et~al.(2016)Zhou, Della, Roberts, Goh, \protect\BIBand{}
  Dhaliwal}]{zhou2016utility}
Zhou H, Della PR, Roberts P, Goh L, Dhaliwal SS (2016) Utility of models to
  predict 28-day or 30-day unplanned hospital readmissions: an updated
  systematic review. \emph{BMJ open} 6(6):e011060.

\bibitem[{Zhu et~al.(2015)Zhu, Luo, Zhang, Shi, \protect\BIBand{}
  Shen}]{zhu2015time}
Zhu T, Luo L, Zhang X, Shi Y, Shen W (2015) Time-series approaches for
  forecasting the number of hospital daily discharged inpatients. \emph{IEEE
  Journal of Biomedical and Health Informatics} 21(2):515--526.

\end{thebibliography}

\end{document}